\documentclass{article}
\usepackage{ijcai22}

\usepackage{times}
\usepackage{soul}
\usepackage{url}
\usepackage[hidelinks]{hyperref}
\usepackage[utf8]{inputenc}
\usepackage[small]{caption}
\usepackage{graphicx}
\usepackage{amsmath}
\usepackage{amsthm}
\usepackage{booktabs}
\usepackage{algorithm}
\usepackage{algorithmic}
\usepackage{amsmath,amssymb,amsfonts}
\usepackage{algorithmic}
\usepackage{graphicx}
\usepackage{textcomp}
\usepackage{xcolor}
\usepackage{booktabs}
\usepackage{color}
\usepackage[normalem]{ulem}
\usepackage{algorithm,algorithmic}
\usepackage{subcaption}

\usepackage{array}
\usepackage{layouts}
\usepackage{lipsum,graphicx,multicol}
\usepackage{multirow}
\usepackage{pgfplots}
\usepackage{tikz}
\usepackage{url} 
\usetikzlibrary{backgrounds}

\usepackage[frozencache ,cachedir=.]{minted}

\setmintedinline{fontsize=\footnotesize,breaklines, breakafter=_}

\def\codesize{\scriptsize}

\pgfplotsset{every axis/.append style={
                    axis x line=middle,    % put the x axis in the middle
                    axis y line=middle,    % put the y axis in the middle
                    axis line style={<->}, % arrows on the axis
                    label style={\small},
                    tick label style={font=\Huge}  
                    }}

\usepackage[labelformat=simple]{subcaption}

\makeatletter
\def\blfootnote{\xdef\@thefnmark{}\@footnotetext}
\makeatother

\newcolumntype{H}{>{\setbox0=\hbox\bgroup}c<{\egroup}@{}}

\newcommand{\myparagraph}[1]{\vspace{-0mm}\paragraph{#1}}

\title{Stochastic Coherence Over Attention Trajectory\\ For Continuous Learning In Video Streams}

\author{
Matteo Tiezzi$^1$
\and
Simone Marullo$^{1, 2}$\and
Lapo Faggi$^{1, 2}$\and
Enrico Meloni$^{1, 2}$ \and \\
Alessandro Betti$^{3}$\and
Stefano Melacci$^1$
\affiliations
$^1$DIISM, University of Siena (Italy) \and
$^2$DINFO, University of Florence (Italy)\\
$^3$ Inria, Lab I3S, MAASAI, Universit\`{e} C\^{o}te d'Azur (France),
\emails
mtiezzi@diism.unisi.it, \{simone.marullo,lapo.faggi\}@unifi.it, meloni@diism.unisi.it, alessandro.betti@inria.fr, mela@diism.unisi.it
}

\begin{document}

\maketitle

\begin{abstract}

% A longstanding goal of Artificial Intelligence is about designing agents that live in an environment and learn by observing the world, with few interactions with the human. From a bare Machine Learning perspective, the challenging nature of this problem is due to the fact that common offline learning procedures, based on large and fully annotated datasets, cannot be exploited. Learning must proceed in an online fashion, data smoothly evolve over time and their order has a precise meaning. Moreover, the capability of interacting with a human calls for a mechanism that allows the agents to focus their attention on precise coordinates of the frame. 
Devising intelligent agents able to live in an environment and learn by observing the surroundings is a longstanding goal of Artificial Intelligence. From a bare Machine Learning perspective, challenges arise when the agent is prevented from leveraging large fully-annotated dataset, but rather the interactions with supervisory signals are sparsely distributed over space and time.
This paper proposes a novel neural-network-based approach to progressively and autonomously develop pixel-wise representations in a video stream. The proposed method is based on a human-like attention mechanism that allows the agent to learn by observing what is moving in the attended locations. Spatio-temporal stochastic coherence along the attention trajectory, paired with a contrastive term, leads to an unsupervised learning criterion that naturally copes with the considered setting. Differently from most existing works, the learned representations are used in open-set class-incremental classification of each frame pixel, relying on few supervisions. 
Our experiments leverage 3D virtual environments and they show that the proposed agents can learn to distinguish objects just by observing the video stream. Inheriting features from state-of-the art models is not as powerful as one might expect. %, opening for further research in the direction of this work.

%Experiments on specifically designed virtual worlds show that the proposed agents can learn to distinguish objects just by observing the video stream, while inheriting features from state-of-the art models is not as powerful as one might expect, opening for further research in the direction of this work.

\end{abstract}

\section{Introduction}
\label{sec:intro}

\blfootnote{Accepted for publication in the 31st International Joint Conference on Artificial Intelligence (IJCAI-ECAI 2022) (DOI: to be annnounced).
}

% AI WOULD LIKE TO CRAFT INTELLIGENT AGENTS
In the context of Artificial Intelligence, the idea of designing \textit{agents that exist in an environment and perceive and act} \cite{10.5555/1671238} is a longstanding goal that introduces a huge number of challenges. %If we consider the role of the visual stimulus,
% COMPUTER VISION + MACHINE LEARNING IS USUALLY ABOUT PRECISE TASKS AND WITH PRETRAINED BACKBONES: MORE RECENTLY, A LOT OF SELF-SUPERVISED LEARNING
While Machine Learning solutions applied to Computer Vision might help in pursuing such a goal, most of their outstanding results are obtained in well-defined vision tasks, leveraging huge collections of supervised data or exploiting pretrained backbones \cite{DBLP:journals/corr/abs-2103-13413}.
More recently, there has been a growing interest in Self-Supervised Learning to learn robust representations without human intervention, still exploiting very large collections of images \cite{jing2020self}. 
% \cite{DBLP:journals/corr/abs-2104-14294}. 
% MOVING FROM THE TASK-RELATED RESULT TO CONTINUAL LEARNING OF AN AGENT: SEVERAL ISSUES!

A lot of issues arise when trying to exploit neural models from other vision tasks in order to design a visual agent that learns while watching a video stream, especially when the agent is expected to parsimoniously interact with humans to get information on what it sees.
%When trying to adapt these neural models to the goal of designing a visual agent that learns by watching a video stream---thus virtually living in an environment that it can observe---and parsimoniously interacting with humans to get information on what it sees, a lot of issues arise. %Amongst them, we mention the lack of a large set of supervised examples, a lot of redundancy in visual information, and the specific data order that cannot be shuffled as in classic stochastic gradient descent. 
Pretrained models might not always help in capturing properties of entities that belong to the particular environment in which the agent lives \cite{Kornblith_2019_CVPR}, and they are subject to inductive biases. 
Moreover, the agent must be able to learn synchronously with the continuous video stream, and the target classes are not known in advance. This setting not only implies redundancy in visual information, but it also introduces constraints in the the data order, that cannot be shuffled as commonly done to implement stochastic gradient descent.
%Moreover, the agent must be able to learn synchronously with the video stream (with redundancy in visual information and the strict constraint that the data order cannot be shuffled, as in classic stochastic gradient descent), and the target classes are not known in advance.
%, and not by randomly sampling data from massive offline collections, as in Self-Supervised Learning \cite{DBLP:journals/corr/abs-2104-14294}. 
%Moreover, the agent must be naturally able to not take a decision  
%on some inputs, since they might belong to unknown categories. 
To this regard, the scientific community is progressively paying more attention to continual learning \cite{parisi2019continual}. %and related problems  \cite{Bendale_2016_CVPR}. %, but these fields are not as mature as the studies on the aforementioned Computer Vision tasks in which Deep Learning succeeds. 
% ATTENTION IS NEEDED!

An often neglected element of crucial importance is the focus of attention, which guides the agent in wild visual scenes and 
attributes precise locations to the human-machine interaction. For example, consider an agent that asks for or receives a specific supervision in a crowded scene, or whenever there is a linguistic interface to exchange information with the human. Without contextualizing the dialogue to what is being precisely observed, the interaction is hardly meaningful. 
In particular, we are referring to the simulation of \textit{human-like visual attention trajectories} \cite{zanca2019gravitational}, which is different from popular neural attention models \cite{chaudhari2019attentive}, that are learnt to cope with a certain task, still relying on large datasets processed offline.
%Attention models in neural networks \cite{chaudhari2019attentive,NIPS2017_3f5ee243} are learned in order to better face a certain task, still relying on large datasets processed offline, and they do not aim at simulating human attention, differently from what is studied at the intersection with Neuroscience \cite{zanca2019gravitational}.
Supervisions can be in the form of a class/instance label about what is being observed, without a precise indication on the boundaries of what is supervised, differently from several Computer Vision tasks \cite{long2015fully}. Fig.~\ref{fig:toy} depicts what we discussed so far (first and second frame).
\iffalse
\begin{figure*}
    \centering
    $\vcenter{\hbox{\includegraphics[width=0.52\textwidth]{fig/toy.pdf}}}$
    \hskip 1cm
    $\vcenter{\hbox{\includegraphics[width=0.24\textwidth]{fig/arch.pdf}}}$    
    \vskip -2mm
    \caption{Left: video stream with static and moving elements. The focus of attention (red circle) disambiguates what is currently under observation, making the human supervision well contextualized (second pic). The agent spatially expands the supervised coordinates to the moving region that includes the focus of attention, learning representations and making predictions out of them (third pic). Right: pixels of frame $V_t$ are encoded into new representations by the neural net $f$. Inference involves all the coordinates (example for $x$), and the classifier $c$ can predict one or more classes or nothing (open-set). Classes are not known in advance. Learning is only about the attended moving area, developing coherent representations over time and space.}
    \label{fig:toy}
    %\vskip -4mm
\end{figure*}
\fi
\begin{figure}
    \centering
    $\vcenter{\hbox{\includegraphics[width=0.45\textwidth]{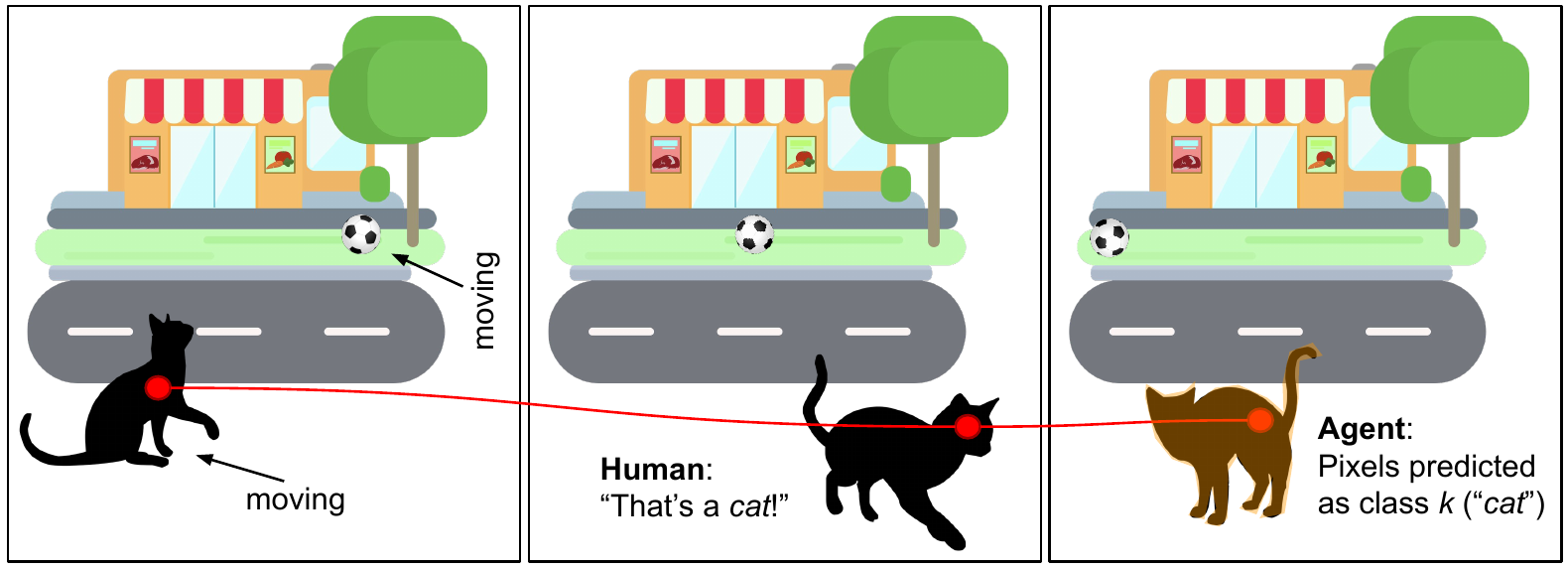}}}$
    \vskip -2mm
    \caption{Stream with static and moving elements. The focus of attention (red circle) disambiguates what is  observed, making the supervision well contextualized (2nd pic). The agent learns (1) pixel-wise representations that are coherent over the focused moving region, and (2) to make predictions (3rd pic). A stochastic sampling procedure yields an \textit{attention graph} that facilitates learning.}
    \label{fig:toy}
    \vskip -2mm
\end{figure}

% *** WHAT WE PROPOSE ***
% - FOA
% - TEMPORAL COHERENCE IN MICRO-SACCADES/SMOOTH PURSUIT
% - SPATIAL COHERENCE BY MOTION
% - CONTRASTIVE TERM
% - STOCHASTIC GRAPH MODELING (differences in neighboring frames + speed)
In this paper, (\textit{i}) we propose a novel approach to online learning from a video stream, that is rooted on the idea of using a scanpath-based focus of attention  mechanism \cite{zanca2019gravitational} to explore the video and to drive the learning dynamics in conjunction with motion information. The attention coordinates offer a precise location for interaction purposes,
%\footnote{Of course, the agent should share the attended location for interaction purposes. Agent attention can be also stimulated by the human, whenever he takes the initiative of providing a supervision.}
and its trajectory has been recently proved to efficiently select the most salient information of the video stream when learning with deep architectures \cite{NEURIPS2020_fc2dc7d2}. 
%Human attention is characterized by fixations, with slower local movements of the gaze (microsaccades, ocular drifts and ocular microtremor, see for example \cite{rucci2015control,RUCCI20161}), and smooth pursuit. Differently, saccades are rapid movements that indicate a switch from a point of interest to another one, exploring new areas that might or might not be about the same visual pattern \cite{zanca2019gravitational}. 
We propose to learn representations that are coherent over the temporal attention trajectory during slow movements of the simulated gaze, that are likely to cover visual patterns with the same semantics. Attention is paired with information coming from motion, that intrinsically suggests the spatial bounds of the attended area. This leads to a spatio-temporal unsupervised criterion that enforces coherence in the representations learned while observing what is moving (Fig.~\ref{fig:toy}). In order to avoid trivial solutions, we augment the criterion with a contrastive term
% \cite{hadsell2006dimensionality} 
that favours the development of different representations in what is inside the moving area and what is right outside of it. %When formalized in a graph-based description, the criterion operates on a spatio-temporal graph whose nodes are pixel coordinates, and edges indicate whether pairs of pixels should be associated to the same representation (or not). 
%However, instead of considering the whole frame, we propose to sub-sample the graph nodes in order to bound and approximately balance the total numbers of the two types of edges. 
Thanks to a graph-based formalization of this approach, 
we define a stochastic procedure that introduces variability in the information provided to the learning algorithm and leads to faster processing, also mitigating the effects of noisy motion information.
% *** WHAT WE PROPOSE *** (CONT.)
% - OPEN SET INCREMENTAL CLASSIFICATION
% - VIRTUAL WORLDS FOR EVALUATION
% - LIST OF CONTRIBUTIONS
 %(\textit{i}) It proposes an approach for online learning in a continuous video stream based on two main pillars, that are human-like focus of attention and spatio-temporal stochastic coherence over the attention trajectory.
(\textit{ii}) We consider the case in which the human intervention is rare, and each supervision is about the coordinates of a single pixel with its class/instance label, thus not a signal that can strongly drive the features development. Target classes are not known in advance, and our model includes an open-set approach to avoid making predictions about unknown elements. 
In order to cope with continual learning, we propose a template-based schema with a dynamic update procedure that is synchronous with the processed stream and efficiently handled by modern hardware. 
(\textit{iii}) Dealing with this setting introduces the further challenge of how to evaluate the artificial agent. We exploit the growing activity in realistic 3D Virtual Environments \cite{DBLP:conf/icpr/MeloniPTGM20},
%that can be easily used to generate photo-realistic video streams with perfect pixel-wise labeling \cite{DBLP:conf/icpr/MeloniPTGM20}, 
designing from scratch (and sharing) three ad-hoc streams with different visual difficulty levels. %The 3D environment is aware of the label of each rendered pixel. 
(\textit{iv}) We compare the learned representations with those from state-of-the art pretrained models that exploited a massive supervision, showing that the latter are not as powerful as one might expect in the considered setting.
% LIST OF CONTRIBUTIONS
%In summary, this paper includes the following list of contributions. (\textit{i}) It proposes an approach for online learning in a continuous video stream based on two main pillars, that are human-like focus of attention and spatio-temporal stochastic coherence over the attention trajectory; (\textit{ii}) When at least one supervision becomes available, the model can make open-set predictions in a class-incremental setting \cite{9040673}, exploiting a template-based technique that dynamically updates supervised data in a synchronous manner; (\textit{iii}) Three benchmarks are created from a recent 3D Virtual Visual Environment for evaluation purposes, comparing the learned representations with features taken from several state-of-the art pretrained models that exploited a massive supervision, showing that they are not as powerful as one might expect in the considered setting, thus opening for further research in the direction of this work.

% ORGANIZATION
%This paper is organized as follows. Section~\ref{sec:related} introduces the related work (in a structured manner). Section~\ref{sec:model} describes the proposed model, that is compared in newly designed benchmarks in Section~\ref{sec:exp}. Finally, Section~\ref{sec:conclusions} draws our conclusions and suggestions for future work.

\section{Related Work}
\label{sec:related}
\iffalse
Main points:
1- Agent
--- Continual Learning (anche (CAL (Alessandro)))
--- Open Set/World Learning, Incremental Learning
--- Zero/few shot learning?
2- Focus of Attention (human-like)
--- Dario/Lapo etc..
--- Focus and MI (NeurIPS last year)
3- Unsupervised learning in vision and invariances
--- Autoencoders (estratto di quello che già Lapo ha cercato)
--- Self-supervised
--- Contrastive learning (minor)
--- Motion/focus coherence in learning over time (CAL (Alessandro))

4- Dense pixelwise predictions => Semantic Labeling
--- State-of-the art semantic labeling models trained in a supervised manner
--- Learning by motion: Facebook learning by watching object moves
5- Virtual worlds (SAILenv)
\fi
\myparagraph{Online, continual, open set learning.} 
%Traditional machine learning techniques usually assume static input data and the existence of a neat distinction between a training and a test phase. Input data, entirely available a the the beginning of the learning procedure, are processed as a whole iterating over the training dataset more epochs in a batch mode fashion. The trained models are then freezed and exploited for inference only and computational expensive re-training procedures are needed to eventually incorporate any new available information. On the other hand, 
An online learning model progressively learns from a stream of data, continuously adapting to every new processed input instance~\cite{hoi2018online}. %, usually resulting in memory and computational efficient algorithms. %The goal of online learning is the same of classical off-line methods, that is to optimize the performance with respect to a given learning task. 
%On a more general level, 
In the specific case of continual learning (also known as life-long, continuous or incremental learning \cite{parisi2019continual}), the goal of the agent is not fixed a priori but changes over time. %, and the agent %is sequentially trained on multiple tasks. The agent 
%should be malleable enough to integrate new knowledge and, at the same time, enough stable to retain old information \cite{abraham2005memory},  %(\textit{stability-plasticity dilemma} \cite{abraham2005memory})
%.  Vanilla neural networks have been shown to struggle in this aspect, since training a network to solve a new task will likely override the information stored in its weights (catastrophic forgetting \cite{mccloskey1989catastrophic,french1999catastrophic}). Three different strategies have been proposed so far to face this problem, that are architectural, regularization and rehearsal strategies
%and different strategies can be adopted for implementation purposes 
%\cite{luo2020appraisal}.
%Recently, physics-inspired approaches have been also proposed  \cite{betti2016principle,betti2019least}.
% WHAT WE PROPOSE VS. ONLINE/CONTINUOUS LEARNING
In this paper, the goal of the agent is to learn to predict the  class labels in every pixel of a video stream. Such goal is not fully defined in advance, since 
%In this paper, the goal of the agent is to learn to predict the categories attached to the pixels of video stream frames, and such goal is not fully defined in advance, since 
the agent becomes aware of classes in function of what the human supervisor tells him, being closer to task-free cases \cite{aljundi2019task}.
%the agent should learn to classify pixels according to the last given supervision and all the old ones and this is done extracting spatio-temporal coherent features over frames that are processed online. % PHYSICS
%The problem of learning over time from a temporal coherent stream of data can also be faced from other perspectives, as for example in the physics-inspired approach of \cite{betti2016principle,betti2019least,achille2018emergence} in which a dynamic evolution of weights is defined through the principle of least action, that is fully compatible with what we propose. 
% OPEN SET
%Our work is also connected to open-set recognition \cite{scheirer2012toward,geng2020recent,bendale2015towards}. %As in typical classifiers, 
Open-set classifiers \cite{scheirer2012toward} can distinguish between examples belonging to different training classes and they can detect whether data do not belong to any of them, that is the case of what we propose. Our work is class-incremental \cite{geng2020recent}, due to the progressive inclusion of new classes after human intervention.\footnote{However, it differs from the protocol of the open-world scenario, that goes beyond what we describe in this paper \cite{geng2020recent}.
One/few shot supervised models \cite{min2021hypercorrelation} also learn new classes from few examples, exploiting prior knowledge.}

%However, it differs from what is referred to as open-world scenario, that is a further extended perspective/protocol that goes beyond what we describe in this paper \cite{geng2020recent}. One/few shot supervised models \cite{wang2020generalizing} also learn new classes from few examples, exploiting prior knowledge.}

%In our setting, pixels not affected by the supervisions given so far have to be classified as unknown, and this must hold until any additional supervision will associate them to some new class. The total number of supervisions is very limited and the spatio-temporal coherence over the attention trajectory plays a key role in extracting consistent features between pixels belonging to the same (moving) object. 
%The majority of pixels is unlabeled so that the proposed framework is also linked to the semi-supervised classification scenario. 

%\vspace{-3mm}
\myparagraph{Focus of attention.} 
%The information our eyes collect is greater than what we are able to process, and the visual system needs a mechanism to locate the most relevant elements in the  scene.
%This is where the Focus Of Attention (FOA) mechanism gets in the game. Visual acuity is indeed restricted to a small portion in the macula region of the retina, the so called fovea. The visual system drives the fovea in the highest informative regions of the visual scene, producing different kinds of eye movements such as fixations, saccades and smooth pursuits. 
Several attempts to model \textit{human-like} focus of attention mechanisms were presented~\cite{borji2012state}, that not only differ in the way they are implemented, but also %from deep learning-based attention models~\cite{borji2019saliency} 
in the nature of the predicted attention (i.e., a temporal trajectory rather than saliency)~\cite{borji2019saliency}. %The majority of the proposed models consider static inputs only and aim to predict the corresponding saliency maps %describing the probability of focusing on each pixel, see for example
%\cite{kummerer2016deepgaze,kruthiventi2017deepfix,cornia2018predicting}. %Then, shifts in visual attention can be indirectly obtained through the winner-take-all and the inhibition of return mechanisms, selecting each time step the most relevant location in space and subsequently decreasing the saliency of the corresponding area~\cite{koch1987shifts}. 
Recently, an unsupervised dynamical model was proposed~\cite{zanca2019gravitational} that
%, differently from most of related approaches \cite{kummerer2016deepgaze,cornia2018predicting}, 
 can be applied both to static images and videos, also studied in the context of online learning in deep networks \cite{NEURIPS2020_fc2dc7d2}---without any loss of generality, it is the model we consider here. %Here the focus of attention is modeled as a point-like particle attracted by a ``mass'' density distribution uniquely determined by low level features in the visual scene (details and motion). Within these models, state of the art results in the scanpath prediction task have been achieved.
%\iffalse
%It was found that viewers tend to fixate the
%center of a selected target object in natural scenes \cite{nuthmann2010object,pajak2013object}, and this still holds in free viewing conditions \cite{stoll2015overt}. Then, considering eye movements during a specific fixation (microsaccades, ocular drifts, ocular microtremor, see for example \cite{rucci2015control,RUCCI20161}), the FOA will likely span the same object, so that the same set of features has to be extracted along the FOA trajectory during the selected fixation. This is also true considering smooth pursuit movements, in which the gaze closely tracks a moving object in the visual scene. Assuming the validity of these results also in case of artificially generated scanpaths, we will enforce spatio-temporal coherence of the extracted features over the attention trajectory. 
%\fi

%\vspace{-3mm}
\myparagraph{Learning invariant features.} 
%Modern deep convolutional neural networks (CNNs) usually consist of several stages of convolutions, activations and pooling layers. At each stage, convolutional features are subsampled by pooling layers, so that top level features attain a local translation invariance. This is of course a desirable property in tasks such as object recognition, in which CNNs achieve state of the art results. An important but often overlooked aspect is that natural images are swamped by nuisance factors such as lighting, viewpoint, part deformation and background. This makes the overall recognition problem much more difficult \cite{lee2011video,anselmi2016unsupervised}
Typical convolutional neural architectures require high sample complexity to learn representations invariant to factors that are not informative for the task at hand. 
%a large amount of data with high variability to learn and generalize well to unseen examples. In other words, the ideal goal is to learn representations that are \textit{invariant} to factors of variation uninformative to the task
%at hand and this is usually achieved, to a small extent, through an high sample complexity.
%Several different but closely related approaches have been proposed so far to deal with such problem \cite{gens2014deep,bergman2019symmetry}, also
%deal with such problems %\cite{gens2014deep}, 
%exploiting information-based criteria \cite{achille2018information}, 
%Several approaches tackle these problems leveraging multiple and temporally coherent views of the same scene \cite{lee2011video}, or exploiting slowness constraints and related ideas \cite{wiskott2002slow}. %, or when tracking objects with a temporal slowness constraint \cite{zou2012deep}.
Some solutions disentangle  
% A large category of approaches rely on the distinction between two different sets of features 
\textit{what} and \textit{where} features, each of them separately encoding informative and uninformative factors of variation \cite{BURT2021145}. 
%For example, \textit{what} features are supposed to encode the semantic content of the input stimulus and to be invariant with respect to its visual variations \cite{hinton2011transforming,denton2017unsupervised}.
%Where features are supposed to codify these variations in an equivariant way instead, that is they have to change according to such transformations. 
%Over the last decade, this latter perspective has been developed through several contributions. 
%These approaches are about static images \cite{hinton2011transforming,zhao2015stacked,leksut2020learning} % and generative models \cite{reed2014learning,makhzani2015adversarial,mathieu2016disentangling}, 
%and videos, distinguishing between content and pose features \cite{denton2017unsupervised,hsieh2018learning} or aiming at disentangling content from motion features \cite{villegas2017decomposing,wang2020g3an,zhu2020s3vae}. 
This paper follows the perspective in which the attention trajectory, paired with motion information, offers a compact way to implicitly constrain the agent to learn invariances. %A lot of common visual transformations %, such as translations, rotations and scale changes, 
%can be indirectly learned by forcing the agent to develop coherent representations during the temporal evolution of the specific attended area.
%Finally, as far as the proposed model is concerned, the goal is to find features that are invariant with respect to the motion of objects in the visual scene and this is achieved requiring their spatio-temporal coherence over the FOA trajectory. This can be considered, to a certain extent, an extension of the aforementioned approaches since lots of common visual transformations such as translations, rotations and scale transformations can be considered as simply arising from the motion of the specific object under consideration. 
%This idea has been recently introduced in \cite{betti2018convolutional,betti2020learning}.
This idea is linked to recent studies about learning invariance to motion in unsupervised learning over time \cite{betti2020learning}.
% FACEBOOK
%also in the case of object foreground
%versus background unsupervised segmentation \cite{}. % One of the principle idea here is that pixels that move together tend to belong to the same object. %The same concept is implemented in our work through the spatial coherence loss. 

%\vspace{-3mm}
\myparagraph{Semantic segmentation.}
Semantic segmentation aims at associating a class label to each pixel of a given image. %In the past few years, astonishing results have been achieved through deep learning models and, in the following, we will briefly review the most well-known techniques. %For more extended surveys, we refer the interested reader to \cite{sultana2020evolution,minaee2021image}. 
Deep architectures for this task usually rely on supervised learning from offline data, including fully convolutional networks \cite{long2015fully}, %to those also exploiting probabilistic graphical models \cite{schwing2015fully,lin2016efficient}.
%One of the most popular deep learning model efficiently performing semantic segmentation was introduced in 2015 and consisted in a network made up of just convolutional layers (fully convolutional network, FCN) \cite{long2015fully}. FCNs disregards the global semantic context of the input images (the empirical receptive field of units at a certain convolutional layer is much smaller than the corresponding theoretical value, especially in higher layers \cite{zhou2014object}) and this makes harder the segmentation task. This issue has been faced so far in several different ways. As far as FCNs are concerned, the authors of \cite{liu2015parsenet} used for example the average feature of the various layers to augment the features in each location. Some other works have exploited probabilistic graphical models in conjunction with CNNs, see \cite{chen2014semantic,schwing2015fully,zheng2015conditional,liu2015semantic,lin2016efficient}. 
%to hourglass-like architectures, encoding the image into a lower-resolution representation and then decoding it into the full-resolution target representations, 
models based on transposed convolutions, dilated convolutions, upsampling and/or unpooling (U/V-net architectures \cite{ronneberger2015u}), 
%Convolutional encoder - decoder architectures are also well established models to perform semantic segmentation, for example we can think of the famous U/V-net architectures \cite{ronneberger2015u,milletari2016v} for biomedical applications, but there exist lots of other works \cite{noh2015learning,badrinarayanan2017segnet,chaurasia2017linknet,cheng2017locality,fu2019stacked,yuan2019object}. Decoders usually exploit deconvolutions (that is, transposed convolutions) and unpooling-layers. One limitation of this approach is the loss of details, discarded during the encoding process, in the final output. 
% feature pyramid networks
% combine multi-scale or different level features,  \cite{zhao2017pyramid}, while 
 transformers \cite{DBLP:journals/corr/abs-2103-13413}.
% applied to semantic segmentation yielded state-of-the-art performance \cite{DBLP:journals/corr/abs-2103-13413,strudel2021segmenter}
% , that are the main ingredients of the DeepLab models \cite{chen2017deeplab,}. 
% Recently,. 
We will compare with these models.

%have made use of feature pyramid networks, that are top-down convolutional networks with lateral connections. 

%To enlarge the receptive field of convolutional layers without increase the total computational cost, \cite{yu2015multi,paszke2016enet} have used dilated convolutions (that is, regular convolutions with up-sampled filters) and this also helps to retain the spatial resolution of the input images. 

%Some other works, as for example \cite{luc2016semantic,xue2018segan}, exploit a generative adversarial framework instead. 

%Finally, there exist also different models that take advantage of some sort of attention mechanism, in conjunction with convolutional networks (see for example \cite{chen2016attention,huang2017semantic,li2018pyramid,fu2019dual}) or in a transformer-like fashion \cite{ranftl2021vision,strudel2021segmenter}. Nevertheless, it is worth noting that these attention mechanisms do not aim at simulating human scanpaths, like in our work instead. For example, the multi-head self attention mechanism in visual transformers \cite{dosovitskiy2020image,ranftl2021vision,strudel2021segmenter} aims to asses the relative importance of patches of pixels at different spatial locations (\textcolor{red}{check this last sentence}). 

%\vspace{-3mm}
\myparagraph{Virtual environments.} The significantly improved quality of the rendered scenes and the intrinsic versatility of 3D Virtual Environments have quickly increased their popularity in the Machine Learning community (e.g., \cite{gan2020threedworld}). 
%Some examples include DeepMind Lab , AI2Thor~\cite{}, ThreeDWorld~\cite{}, HabitatSim~\cite{savva2019habitat}, iGibson~\cite{xia2020interactive}. 
%The quality of the graphics depends both on the 3D engine on which the environments are built and on the level of detail of the designed scenes. 
%In general, these environments come interfaces to high-level programming languages. 
SAILenv~\cite{DBLP:conf/icpr/MeloniPTGM20} is a recently proposed environment based on Unity3D, specifically aimed at easy customization and interface to Machine Learning libraries. SAILenv yields pixel-level motion information from Unity3D, and provides utilities to ease the generation of ad-hoc scenes for continual learning scenarios~\cite{meloni2021evaluating}, making it well suited for what we propose.

%These environments are designed to be interfaced with high-level programming languages and, in turn, with common Machine Learning libraries. 

%Several different tasks are studied using these 3D simulators, such as generic robot navigation, visual recognition, visual QA -- see \cite{Duan2021} and references therein. 

%Most virtual environments allow for basic agent navigation in closed door environments and limited physical interaction, while some also have photo-realistic and moving objects. 

%It not surprising to see 3D Virtual Environments dedicated to Machine Learning and, more generally, AI that are built within 3D game engines, designed to render high-quality graphics at large frame rates. 

%The visual quality of the rendered scenes depends both on the features of the 3D engine and on the design of the 3D models that are shared together with the environment itself. 

%Some environments are developed in the context of open projects that might benefit from the contributions of large communities \cite{deitke2020robothor,kolve2017ai2,weihs2020allenact}, while others are based on closed source solutions \cite{gan2020threedworld}.

\section{Model}
\label{sec:model}
\begin{figure}[t!]
    \centering
    \scriptsize
    \includegraphics[width=0.475\textwidth]{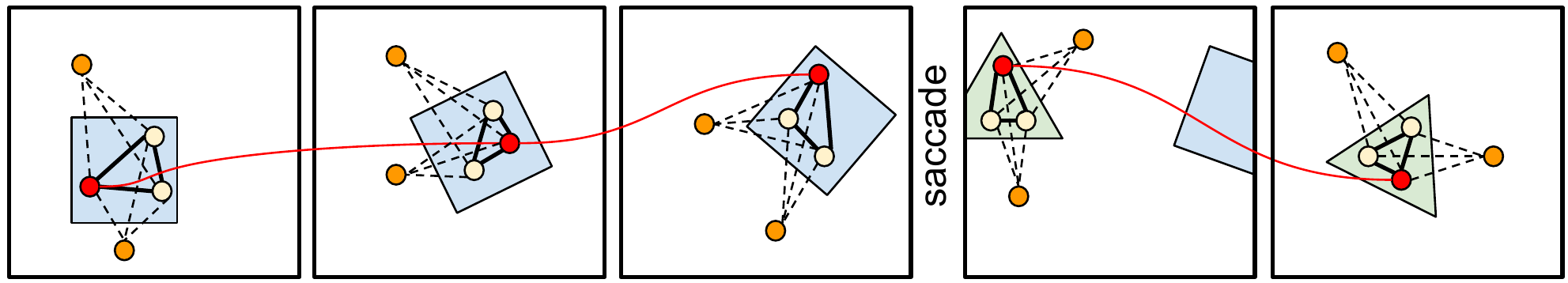}\\
    \vskip -1mm
    $t=0$\hskip 1.1cm$t=1$\hskip 1.1cm$t=2$\hskip 1.1cm$t=3$\hskip 1.1cm$t=4$
    \vskip -1mm
    \caption{A rectangle rotates toward the right, attracting the attention $a_t$ (red nodes) that explores it. Then, a triangle enters the scene, and the simulated gaze quickly moves on it (saccade). The red path links nodes on which we enforce \textit{temporal coherence} (not in saccades). %Stochastic attention graph: 
    For each $t$, some coordinates (yellow) are sampled in the \textit{moving} region that includes $a_t$, while 
    other points (orange) are sampled outside of it. Solid lines (positive edges) link \textit{spatially coherent} samples; dashed  lines (negative edges) are about the \textit{contrastive} term.}
    \label{fig:graph}
    \vskip -4mm
\end{figure}

\begin{figure}
    \centering
    $\vcenter{\hbox{\includegraphics[width=0.44\textwidth]{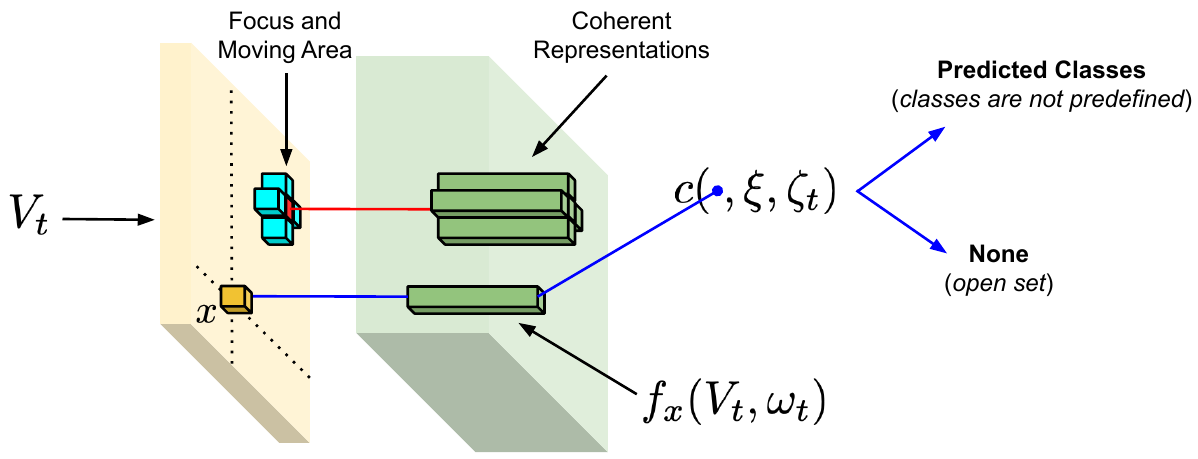}}}$    
    \vskip -2mm
    \caption{Pixels of frame $V_t$ are encoded into new representations by the neural net $f$. Inference involves all the coordinates (example for $x$), and the classifier $c$ can predict one or more classes or nothing (open-set). Classes are not known in advance. Learning is only about the attended moving area, developing coherent representations over time and space.}
    \label{fig:toya}
    \vskip -2mm
\end{figure}

The basic concepts of the proposed unsupervised feature extractor are summarized in the example of Fig.~\ref{fig:graph}. Our method is based on the assumption that a \textit{human-like attention} trajectory \cite{zanca2019gravitational} generally spans the important location of the stream, and the gaze moves more slowly within areas with uniform semantic properties, whose bounds can be further guessed by motion information. 
%areas with uniform semantic properties \cite{zanca2019gravitational}, and motion naturally provides additional information to guess. 
Hence, we enforce both \textit{temporal} and \textit{spatial coherence} constraints to force pixel embeddings to be consistent in time and space, with respect to locations virtually connected by the attention trajectory and by motion, respectively. We exploit a \textit{contrastive loss} to avoid trivial solutions, and a \textit{stochastic approach} to lighten the  computational burden. Given the pixel embeddings, a template-based classifier learns how to classify each of them in an open-set class-incremental setting (Fig.~\ref{fig:toya}).

% VIDEO STREAM, NETWORK, ONLINE LEARNING
Let us indicate with $X$ the set $\mathbb R^{w\times h\times c}$, and consider a continuous and potentially life-long video stream $V\in(X)^\mathbb N$ that, at the discrete time index  $t \in \mathbb N\mapsto V_t\in X$, yields the video frame $V_t$ at the resolution of $w \times h$ pixels with $c$ channels. 
%Without any loss of generality, we represent $V_t$ by a multidimensional array in $X = \mathbb{R}^{w \times h \times c}$.
We also consider a neural network model that implements the function $f(\cdot, \omega)\colon X \rightarrow F$, with $F = \mathbb{R}^{w \times h \times d}$, where $\omega$ indicates the weights and biases of the net. The network is designed to process a frame $V_t$, and $f(V_t, \omega)$ is an encoded representation of the frame where, for each pixel, we have a vector with $d$ components (Fig.~\ref{fig:toya}). The network evolves over time and $\omega_t$ are the weights and biases at time $t$. We denote with $f_x$ the output of $f$ restricted to the pixel at 2D coordinates $x \in Z^\diamond: = \{1,\dots,w\} \times \{1,\dots,h\}$. In online learning from a video stream $V$, the network weights $\omega_{t+1}$ are obtained updating the previous $\omega_{t}$ with a law that depends on the gradient of a suitable loss function $L$ with respect to $\omega$.

% FIG 3 ERA QUI [STEFANO] FIG DELLA RETE

% FOCUS OF ATTENTION
%\vspace{-3mm}
\myparagraph{Human-like attention.} 
The first pillar sustaining our model is a function 
$t\mapsto a(t)$, yielding an explicit estimate of the 2D coordinates where human would focus the attention at each time $t$.
% \footnote{To introduce the following differential equations as in related work, here $t$ is a continuous parameter.} 
Among the various attention-prediction models yielding temporal trajectories \cite{borji2012state}, the recent unsupervised model \cite{zanca2019gravitational} achieved state-of-the art results in simulating human-like attention. The authors showed that visual cues of the frame can act as gravitational masses. The equation of the potential of the gravitational field,
$\varphi(x,t) :=-{{(2\pi)}^{-1}}\int_{Z}\log\Vert x-z\Vert\mu(z,t)\,dz$,
is at the basis of the attention model, where $Z$ is the continuous set of frame coordinates and $\mu(x,t)$ is the total mass at $(x,t)$. 
% The authors in \cite{zanca2019gravitational} showed that the magnitude of the brightness and an optical-flow-based measure of motion activity
% can be modeled as masses yielding state-of-the art results in simulating human-like attention.
In particular, the magnitude of the brightness and an optical-flow-based measure of motion activity can be modeled as masses, and their impact is controlled by two positive scalars $\alpha_b$ and $\alpha_m$, respectively, and an inhibitory signal is also included.
%In order to compute $a(t)$, the following differential equation
%with initial conditions $a(0)=a^0$ and $\dot a(0)=a^1$
%is integrated,
The  focus of attention is modeled as a point-like particle subject to the 
 above potential, and its trajectory  $a(t)$ is determined integrating the following  equation with initial conditions $a(0)=a^0$ and $\dot a(0)=a^1$,
\begin{equation}
\ddot a(t) + \rho \dot a(t) +\nabla\varphi(a(t),t) = 0 \qquad t>0,
\label{eq:geymol}
\end{equation}
where the dissipation is controlled by $\rho>0$
and $\nabla\varphi$ is the spatial gradient of the potential. 
% This model is not based on pre-computed saliency maps, and the resulting trajectories include gaze movements with human-like patterns. 
The gaze performs \textit{fixations} in locations of interest, with relatively low speed movements.
% in the fixation area.
\textit{Smooth pursuit} consists of slow tracking movements performed to follow a considered stimulus. Differently, \textit{saccades} are fast movements to relocate the attention toward a new target. In the most naive case, the first two categories of patterns can be distinguished by the latter exploiting a thresholding procedure on velocity, based on $\nu > 0$.

% TEMPORAL COHERENCE LOSS
%\vspace{-3mm}
\myparagraph{Temporal coherence.} 
During fixations, the attention spans a certain part of an object, with limited displacements of the gaze. Similarly, during smooth-pursuit the attended moving area has uniform semantic properties.
% While performing fixations,  and smooth-pursuit, the attention is localized in areas with uniform semantic properties
% Fixation usually involves a certain part of an object, with limited displacements of the gaze over such part. Similarly, attention could simply track a moving object.
% Our model is built on the assumption that while performing fixations and smooth-pursuit, the attention is localized in an area with uniform semantic properties.
% A fixation usually involves a certain part of an object, with limited displacements of the gaze over such part. Similarly, attention could simply track a moving object.
Differently, during saccades, the attention switches the local context, shifting toward something that might or might not belong to the same object.
We implement the notion of \textit{temporal coherence} defining the loss function $L_T$ that ($i.$) restricts learning to the attention trajectory, filtering out the information in the visual scene \cite{NEURIPS2020_fc2dc7d2} and ($ii.$) avoids abrupt changes in the feature representation during fixations and smooth pursuit,
\begin{equation}
    L_{T}(\omega, \hat{\omega}, t)  := \delta_t \left\|  f_{a_t}(V_{t}, \omega) - f_{a_{t-1}}(V_{t-1}, \hat{\omega}) \right\|^2,
    \label{eq:temporal_coherence}
\end{equation}
%\[\A{L_T(\omega,v,t):=\frac{\delta_t}{2}\Vert f_{a_t}(V_{t}, \omega)- f_{a_{t-1}}(V_{t-1},\hat\omega)\Vert^2}\]
where $\|\cdot\|$ is the Euclidean norm,
and $\delta_t$ is equal to $0$ in case of saccades, otherwise it is $1$.\footnote{We also consider the case of feature vectors with unitary Euclidean norm---see supplementary material.
}
% \footnote{We also consider the case of feature vectors with unitary Euclidean norm. In this case, $L_T$ and the following losses are redefined using the dot product---see supplementary material.
When plugged into the online optimization, $\omega\gets\omega_t$ and $\hat{\omega}\gets\omega_{t-1}$. This loss is forced to zero for $t=0$, and it will be paired with other penalties described in the following, thus avoiding trivial solutions.

% MOTION AND SPATIAL ATTENTION GRAPH
\myparagraph{Spatial coherence.} Temporal coherence is not enough to capture the spatial extension of the encoded data, since it is only limited to the attention trajectory. Motion naturally provides such spatial information, and we follow this intuition designing agents that learn by observing moving attended regions.\footnote{Filtering out camera motion, e.g., in agents aware of how the camera moves (devices with sensors) or using software techniques.
%\cite{Ranjan_2019_CVPR}.
} As a matter of fact, motion plays a twofold role, being crucial in defining the attention masses and as a mean to extend the notion of coherence to a frame region, i.e., \textit{spatial coherence}.
Formally, for each fixed $t\in\mathbb N$, we indicate with $S_t \subseteq Z^\diamond$ the set of frame coordinates that belong to the region of connected moving pixels that includes $a_t$.\footnote{Moving pixels are selected as detailed in the suppl. material.}
We introduce what we refer to as \textit{spatial attention graph} at $t$, $\mathcal G_t$, with a node for each pixel of the frame ($\forall x\in Z^\diamond$) and with two types of edges, referred to as positive and negative edges. Positive edges link pairs of nodes whose coordinates belong to $S_t$, while negative edges link nodes of $S_t$ to nodes outside the moving region.
% SPATIAL COHERENCE LOSS
The positive edges of the attention graph allow us to introduce a spatial coherence loss $L_{S}$,
\begin{equation}
    L_{S}(\omega, t)  := \frac{1}{2} \hskip -5mm \sum_{\hskip 5mm \substack{x,z \in S_t, x \neq z}} \hskip -5mm \left\| f_{x}(V_{t}, \omega) - f_{z}(V_{t}, \omega) \right\|^2,
    \label{eq:spatial_coherence}
\end{equation}
that encourages the agent to develop similar representations inside the attended moving area $S_t$, while $\omega \gets \omega_t$. The notion of learning driven by the fulfilment of spatio-temporal coherence over the attention trajectory ($L_{S}$ and $L_{T}$ of Eq.~\ref{eq:temporal_coherence} and Eq.~\ref{eq:spatial_coherence}) is the second pillar on which our model is built. %, since Eq.~\ref{eq:temporal_coherence} bridges representation at consecutive time instants by means of the attention coordinates, while Eq.~\ref{eq:spatial_coherence} expands coherence over the moving region  (recall that $a_t$ belongs to $S_t$).

% CONTRASTIVE TERM
\myparagraph{Contrastive loss.}
% The learning problem driven by spatio-temporal coherence only is not well-posed, since a constant solution trivially solves it. We propose t
In order to prevent the development of trivial constant solutions, which fulfill the spatio-temporal coherence, we add a \textit{contrastive} loss $L_C$ that works the opposite way $L_S$ does. In particular, $L_C$ exploits negative edges of $\mathcal{G}$ to foster different representations between what is inside the moving area and what is outside of it,
\begin{equation}
    L_{C}(\omega, t)  := {\left( \hskip -5mm \sum_{\hskip 5mm x \in S_t,z \in O_t} \hskip -7mm \left\| f_{x}(V_{t}, \omega) - f_{z}(V_{t}, \omega) \right\|^2 + \hskip -0.5mm \varepsilon \right)^{\hspace{-2mm}-1}},
    \label{eq:contrastive}
\end{equation}
where $O_t = Z^\diamond \setminus S_t$ is composed of frame coordinates not in $S_t$ and $\varepsilon > 0$ avoids divisions by zero. We notice that this contrastive loss is different from InfoNCE \cite{oord2018representation}.

% STOCHASTIC APPROXIMATION
\myparagraph{Stochastic coherence.} These spatial and contrastive losses are plagued by two major issues. First, the number of pairs in Eq.~\ref{eq:spatial_coherence} and Eq.~\ref{eq:contrastive} is large, being it $(1/2) |S_t|(|S_t|-1)$ and $|S_t|(wh -|S_t|)$, respectively, making the computation of the loss terms pretty cumbersome. 
Secondly, whenever an exact copy of a moving object appears in a not-moving part of the scene, there will be pixels of the first instance that are fostered to develop different representations with respect to pixels of the second one, what we refer to as collision. However, this clashes with the idea of developing a common representation of the object pixels.
%Secondly, whenever the exact same object instance appears twice in the same frame, moving and not-moving, the contrastive loss might try to enforce incoherence on pairs where spatial coherence is also enforced, that is what we refer to as collision. 
Hence, we replace $\mathcal{G}_t$ with the subgraph $\tilde{\mathcal{G}}_t$, that is the \textit{stochastic spatial attention graph} at time $t$, composed by nodes that are the outcome of a stochastic sub-sampling of those belonging to $\mathcal{G}_t$. In particular, the node set of $\tilde{\mathcal{G}}_t$ is the union of $\tilde{S}_t \subseteq S_t$ and $\tilde{O}_t \subseteq O_t$, where $a_t$ is guaranteed to belong to the former. Edges of $\tilde{\mathcal{G}}_t$ are the positive and negative edges of $\mathcal{G}_t$ connecting the subsampled nodes. 
% SAMPLING
The key property of the stochastic graph is that the number of positive and negative edges is a chosen $e > 0$.\footnote{Once we select $e$, if $s:=|\tilde{S}_t|$ and $o:=|\tilde{O}_t|$, we have $e = s(s - 1)/2$ positive edges and $e = s o$ negative ones.
% , and we can solve for $s$ and $o$. 
We approximated the solution with $s = \left\lfloor(1 + \sqrt{1 + 8e}\,)/2 \right\rfloor$ and $o = \left\lceil e/s\right\rceil$.}
The set $\tilde{S}_t$ is populated by uniformly sampling nodes in $S_t$, ensuring that $a_t$ is always present, while $\tilde{O}_t$ is populated by sampling from a Gaussian distribution centered in $a_t$ and with variance $\sigma$, discarding samples $\notin O_t$.\footnote{We selected $\sigma$ to be $\propto \sqrt{|S_t|}$ by means of an integer spread factor $\beta \geq 1$. We repeat the Gaussian sampling until we collect the target number of points in $\tilde{O}_t$, up to a max number of iterations.} Large $\sigma$'s lead to sampling data also far away from the focus of attention, while small $\sigma$'s will generate samples close to the boundary of the moving region. 
In the example of Fig.~\ref{fig:graph} we emphasized how the attention bridges multiple instances of $\tilde{\mathcal{G}}_t$ over time, yielding a \textit{stochastic attention graph}.
% BENEFITS OF STOCHASTIC ATTENTION GRAPH
Such a graph reduces the probability of collisions, both due to the random sampling and to the control on the sampled area by means of $\sigma$, and it introduces variability in the loss functions also when computed on consecutive frames. Moreover, the impact of imperfect segmentation of the moving region is reduced, since only some pixels are actually exploited, re-sampled at every frame. Since the number of pairs is bounded, the stochastic graph makes the formulation suitable for real-time processing.

% FIG ERA QUI [STEFANO] fava

% FINAL LOSS AND UPDATE RULE FOR ONLINE LEARNING
\vspace{-0.5mm}\myparagraph{Cumulative loss.} We define $L$ as the cumulative loss at a certain time instant $t$, where the contributes of Eq.~\ref{eq:temporal_coherence}, Eq.~\ref{eq:spatial_coherence}, Eq.~\ref{eq:contrastive} are weighted by the positive scalars $\lambda_T$, $\lambda_S$ and $\lambda_C$,
\begin{equation}
    L(\omega,\hat \omega, t):=\lambda_TL_T(\omega,\hat\omega,t)
+ \lambda_S\tilde L_S(\omega,t)+
\lambda_C \tilde L_C(\omega,t).
\end{equation}
The losses $\tilde L_S$ and $\tilde L_C$ are the stochastic counterparts of $L_S$ and $L_C$, respectively, in which the sets $\tilde S_t$ and $\tilde O_t$ are used in place of $S_t$ and $O_t$.
We define $\nabla L$ to be the gradient of 
the loss with respect to its first argument, that drives the online learning process,
$\omega_0=\varpi,\ \omega_{t+1}=\omega_t-\alpha\nabla L(\omega_t,\omega_{t-1},t),\  t\in\mathbb N$, 
with $\alpha > 0$ (learning rate) and $\varpi$ some random initialization of the parameters of the network. We remark that other recent learning schemes for temporal domains could be exploited as well \cite{NEURIPS2020_fc2dc7d2}. 

% SCENARIO: OPEN SET CLASS INCREMENTAL CLASSIFICATION
\vspace{-0.5mm}\myparagraph{Pixel-wise classification.} A human supervisor occasionally provides a supervision at coordinates $a_t$ about a certain class $y_t$.\footnote{
%For the purpose of this paper, we do not deepen into how natural language can be translated into a class index, that is another huge research field \cite{mukerjee2006grounded}. It is enough to think that each class is mentioned with the same linguistic label by the human, with no ambiguities, so that it can be translated into a numerical class index. Moreover, 
Notice that only one pixel gets a supervision a at time $t$, thus it is different from few-shot semantic segmentation \cite{min2021hypercorrelation}.} Let us define an open-set classifier $c(\cdot, \xi, \zeta)\colon F \mapsto Q$ that predicts the class-membership scores (belonging to a generic set indicated with $Q$), over a certain number of classes that can be attached to each pixel-level representation. 
The main parameters of the classifier are collected in $\zeta$, and when all the membership scores are below threshold $\xi$ the classifier assumes to be in front on an unknown visual element and it does not provide a decision (\textit{open-set})---see Fig.~\ref{fig:toya}.
Whenever a supervision on a never-seen-before class is received,
the classifier becomes capable of computing the membership score of such class for all the following time steps (\textit{class-incremental}). %\footnote{The meaning of each supervision depends on the supervisor, that might intend to operate at a category or instance level.} 
%As a result, the number of classes is a monotonically increasing function of time. %, so that $q_t \geq 0$ membership scores are produced at $t$, $q_0 = 0$. 
We consider the case in which supervisions are extremely rare, not offering a suitable basis for gradient-based learning of $c$ or further refinements of $f$. The previously described unsupervised learning process is what is crucial  to learn compact representations that can be easily classified, since it favours pixels of the same object to be represented in similar ways over time and space. %\footnote{%We force coherence for the instance that is moving at the considered time. %, with no guarantees that different instances of the same category will be represented by the same features. 
%Of course, the same object could or could not be represented with the same fethe classifier $c$ can predict the same class on different feature vectors.}

% NEAREST NEIGHBOR CLASSIFIER AND DYNAMIC UPDATE OF TEMPLATES
The most straightforward way to implement the  open-set $c$ is with a distance-based model, storing the feature vectors associated to the supervised pixels as templates.\footnote{We tested the squared Euclidean distance and cosine similarity.} This allows the model to not make predictions when the minimum distance from all the templates is greater than $\xi$. We indicate with $(k_t, y_t)$ a supervised pair where $k_t = f_{a_t}(V_t, \omega_t)$ is the template at coordinates $a_t$ of frame $V_t$. The intrinsic dependence of $k_{t}$ on the time index could make templates become outdated during the agent life, for example due to the evolution of the system, so that for some $t'>t$ we might have $\| k_t - f_{a_t}(V_t, \omega_{t'})\| \gg 0$, leading to potentially wrong predictions. 
In order to solve this issue, we propose to dynamically update the templates. Modern hardware and software libraries for Machine Learning are designed to efficiently exploit batched computations %, and neural models efficiently handle batched  inputs. 
In online learning, this feature is commonly not exploited, since only one data sample becomes available for each $t$. We indicate with $B_t$ a special type of mini-batch of frames, $B_t = \{V_t\} \cup \{V_r,\ r \in H_t \}$, composed of the current $V_t$ and the frames associated to supervised time instants whose indices are in $H_{t}$. Due to the tiny number of supervisions, storing supervised frames does not introduce any critical issues, and a maximum size $b > 1$ for each mini-batch can also be defined, populating $B_t$ with up to $b-1$ previously supervised frames, chosen with or without priority. This way, batched computations can be efficiently exploited to keep templates up-to-date.
\vspace{-1mm}

% TO BE DISCUSSED IN THE EXPERIMENTAL SECTION, SOMEWHERE
% Normalization (refer to the supplementary material - appendix A)
% How the moving segment is built (supplementary material - appendix B)
% Emphasize the fact the we assume to have a fixed camera but we could also remove this assumption and filter out the camera movement as a preprocessing
% Benchmark description - using Virtual Environment (as a proposed element) - prepare appendix C to say more about them and about the video data we are sharing.
% Say that we consider the case of mutually exclusive classes, even if our formulation is generic
% Say what are the parameters of the FOA and the initial conditions $a^{0}$ and $a^{1}$

\section{Experiments}
\label{sec:exp}
In order to create the right setting to evaluate what we propose, we need continuous streams able to provide pixel labels at spatio-temporal coordinates that are not defined in advance. % since they depend, for example, on the evolution of the attention trajectory and other design choices. It is also crucial to have the possibility of controlling the properties of the stream to investigate how our approach behave in well-defined (and reproducible) conditions.
%The empirical evaluation of agents that learn from continuous streams is challenging  due, for example, to the cost of setting up dense frame-wise supervisions. 
% Going beyond the challenging design of agents that learn from continuous streams, further difficulties arise from the empirical evaluation of their performance,  since it is extremely expensive to setup full-frame supervisions in very long streams.
% Related benchmarks available in literature are either usually composed of short-length (hundreds of frames) videos or of larger-scale partially-labeled videos \cite{yu2018bdd100k}, since it is extremely expensive to setup full-frame supervisions in very long streams. The attention could explore areas where the ground truth is missing in popular datasets.

%Indeed, in this context the capability to process potentially lifelong videos, the availability of pixel-wise supervisions on the whole frame for performance evaluation or the management of the complexity of the visual scene at hand could represent important factors to foster an all-around evaluation.
%\vspace{-3mm}
\vspace{-1mm}\myparagraph{Virtual environments.}
%The empirical evaluation of agents that learn from continuous streams is challenging  due, for example, to the cost of setting up dense frame-wise supervisions. 
We consider photo-realistic 3D Virtual Environments within the  SAILenv platform \cite{DBLP:conf/icpr/MeloniPTGM20}, that include pixel-wise semantic labeling and motion vectors of potentially endless streams,
% The  provides access to pixel-level labeling as long as motion vectors inherited from the underlying 3D engine. 
creating (from scratch) three 3D scenes to emphasize different and challenging aspects on which we measure the skills of the agents.\footnote{
% All that is needed to run experiments, including code (PyTorch--Modified BSD license)
Code, data and selected hyper-parameters can be downloaded at \url{https://github.com/sailab-code/cl_stochastic_coherence}}. 
% The agent observes the scene from a fixed location, and some objects of interest move along pre-designed smooth trajectories while rotating and getting closer to/farther from the camera.
% Objects move one at a time, and the next object moves when the previous one completes its route going back to its original location, what we refer to as a \textit{lap}.
The agent observes the scene from a fixed location, and some objects of interest move, one at the time,  along pre-designed smooth trajectories while rotating and getting closer to/farther from the camera. We denote with the term \textit{lap} a complete route traveled by each object to its starting location.

\begin{figure}
    \centering
    %\begin{minipage}{0.48\textwidth}
    \fbox{\includegraphics[width=.13\textwidth]{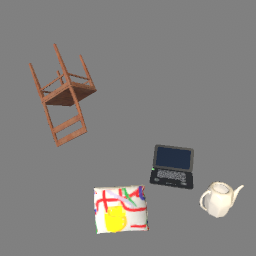}}
    %\hskip 1mm 
    \fbox{\includegraphics[width=.13\textwidth]{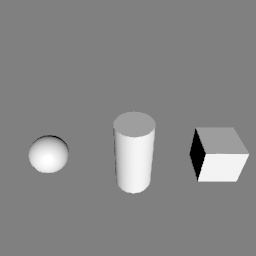}}
    %\hskip 1mm 
    \fbox{\includegraphics[width=.13\textwidth]{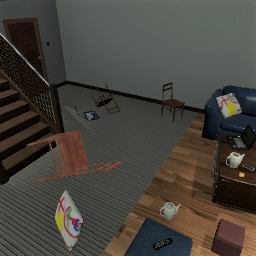}}\\
    %\vskip -3.5mm
    \tiny
    \hskip 0.0cm \textsc{EmptySpace} \hskip 1.4cm \textsc{Solid} \hskip 1.6cm  \textsc{LivingRoom}\\
    %\end{minipage}
    %\hskip 1mm
    %\begin{minipage}{0.44\textwidth}
    %\centering
    %\fbox{\includegraphics[width=.1\textwidth]{fig/sampling/original_pairs_1000_spread_05.pdf}}
    \vskip 1mm
    \fbox{\includegraphics[width=0.1\textwidth]{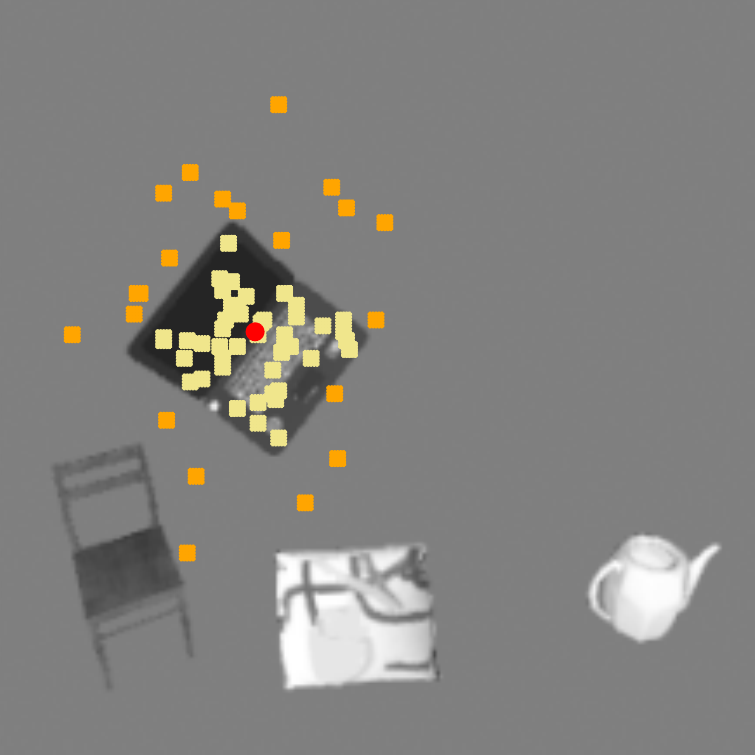}}
    %\fbox{\includegraphics[width=.24\textwidth]{fig/sampling/plot_sampling_pairs_1000_spread_5.pdf}}
    %\fbox{\includegraphics[width=.1\textwidth]{fig/sampling/original_pairs_20000_spread_1.pdf}}
    \fbox{\includegraphics[width=0.1\textwidth]{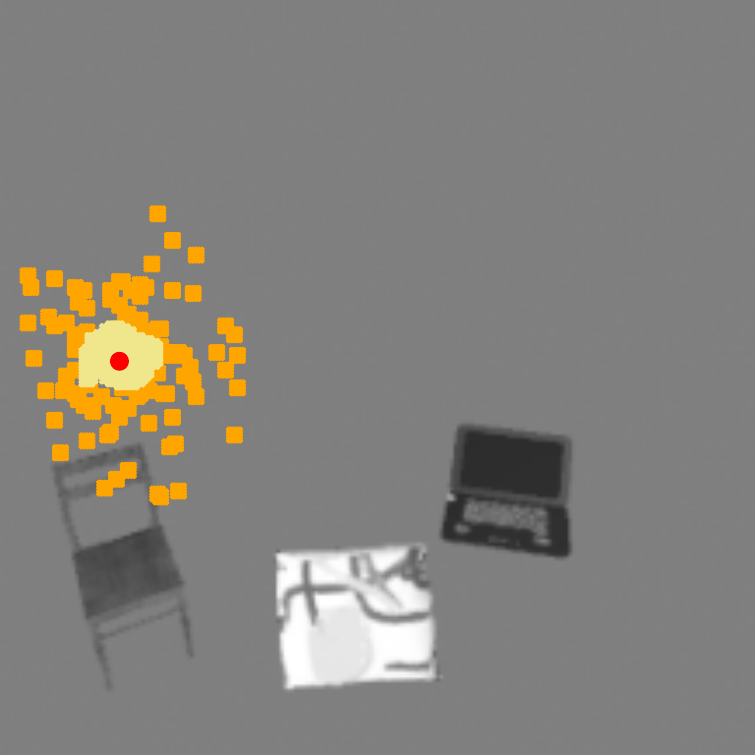}}
    \fbox{\includegraphics[width=0.1\textwidth]{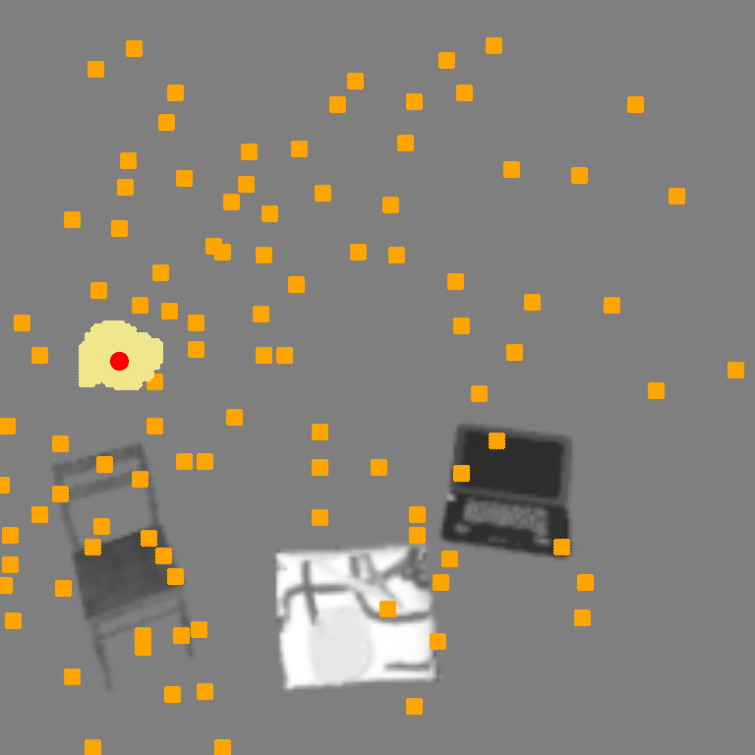}}\\
    %\vskip -3.5mm
    \tiny
    \hskip 0.1cm $e=1\,$k, $\beta=1$ \hskip 0.3cm $e=20\,$k, $\beta=1$ \hskip 0.3cm $e=20\,$k, $\beta=5$\\
    \vskip -1mm
    %\end{minipage}
    \caption{Top: sample frames from the three 3D scenes/streams (objects rotate and scale while moving). Bottom: samples of stochastic spatial graphs in \textsc{EmptySpace} (same color patterns of Fig.~\ref{fig:graph}).}
    \label{fig:streams}
    \vskip -4mm
\end{figure}

%Interestingly, SAILenv produces the visual stream complemented by motion-related pixel-wise information on the whole scene as long as complete semantic labeling, coming in handy for our purposes.
We designed three different scenes. (\textit{i})
\textsc{EmptySpace}: four photo-realistic textured object from the SAILenv library (chair, laptop, pillow, ewer) move over a uniform background. The goal is to distinguish them in a non-ambiguous setting.
%(1650 frames)
(\textit{ii}) \textsc{Solid}: a gray-scale environment 
%composed by 400 frames 
 with three white solids (cube, cylinder, sphere) is considered. Due to the lack of color-related features, %it is hard to discern the solids unless 
 the agent must necessarily develop the capability of encoding information from larger contexts around each pixel. 
(\textit{iii}) \textsc{LivingRoom}:
%(658 frames)
 the objects from \textsc{EmptySpace} are placed in a photo-realistic living room composed by other non-target objects (i.e., an heterogeneous background with a couch, tables, staircase, door, floor), and multiple static instances of the objects of interest.
%Hence, we start from grayscale 3D-geometric shapes in a uniform background up to a photorealistic living room populated by some real-life object of interest with complex moving patterns. 
%We restrict our analysis to the case  of mutually exclusive classes, even if our formulation is generic. 
Samples of are shown in Fig.~\ref{fig:streams} (top) and, in what follows, $n$ is the number of target objects in each stream, while $m=n+1$ is the total number of categories (including the ``unknown'' class). %(i.e., adding the category that includes everything that is not one of the $n$ objects---that is the one that we expect to be predicted as ``unknown'' in our experiments).
% forse meglio rimuovere numero di frames da qui 

\setlength{\fboxsep}{0pt}%
\setlength{\fboxrule}{1pt}%

%\vspace{-3mm}
\vspace{-1mm}\myparagraph{Setup.} We created three pre-rendered 2D visual streams by observing the moving scenes ($256 \times 256$ pixels, with $\approx 51\,$k, $12\,$k and $20\,$k frames,  respectively, corresponding to $31$ completed laps for each object), both in grayscale (\textsc{BW}) and color (\textsc{RGB})---\textsc{Solid} is \textsc{BW} only.
% To accelerate the experimental validation we generated $\approx 51\,$k, $12\,$k and $20\,$k pre-rendered frames from the three aforementioned streams, respectively, that correspond to $31$ completed laps for each object.
%we extrapolated from each stream a video, which was repeated 25 times (yielding 10k, 41k and  16k frames-long videos, for \textsc{Solid}, \textsc{EmptySpace}, \textsc{LivingRoom} respectively)  in order to constitute the learning data on which the unsupervised criterion of stochastic coherence learning described in Section \ref{sec:model} is performed\footnote{Note that we employ motion-related information (optical flow) directly produced from SAILenv. This statistics could be obtained, in real-life videos, from other algorithms \cite{horn1981determining}. Moreover, we assume to deal with the fixed-camera scenario to ease the optical-flow computation. This assumption can be easily removed filtering out the camera movement as a preprocessing step.}. 
The agent learns by watching the first $25$ laps per object and, only in the subsequent laps, receives a total of $3$ supervisions $(k_t, y_t)$ per object, spaced out by at least $100$ frames.
% time instants starting from the first frame of the route and waiting at least $100$ frames before supervising  the same object again (of course, $a_t$ must belong to the supervised object, and we set $b = 1+3n$ to ensure a frequent update of the templates).
Learning stops when all the objects complete $30$ laps and, finally, performances are measured in the the last lap, considering the F1 score (averaged over the $m$ categories), either along the attention trajectory or in the whole frame area.
%Afterwards, 5 repetitions of the video are used to provide and update the supervised pairs $(k_t, y_t)$ at the FOA coordinates $a_t$ (only when the attention coordinates slides inside the area of an object of interest). We provide 3 supervisions for each object class. 
%We set the maximum size $b$ of the frames mini-batch, $B_t$, to be $b = |classes| * 3$, with $H_t$ containing the indices of every 100 frames. 
% detto malissimo - vorrei dire che la supervisione viene data ogni 100 frames 
%This task is automated employing the full-frame labels provided by the Virtual Environment.
%The performance of the developed models is evaluated on a temporal window consisting of a single repetition of the video. We report in the results of this section the F1 measure on the FOA trajectory and over the whole-frame, computed over the whole temporal window and averaged over three runs. 
We evaluated the proposed approach considering two different families of deep convolutional architectures yielding $d$ output features, referred to as \textsc{HourGlass} ( UNet-like \cite{ronneberger2015u}) and \textsc{FCN-ND} (6-layer Fully-Convolutional without any downsamplings \cite{sherrah2016fully}), respectively (see the supplementary material for all the details).
%We denote with \textsc{HourGlass} a UNet \cite{ronneberger2015u} inspired architecture\footnote{The implementation is available at  \url{https://github.com/usuyama/pytorch-unet}} where we explored several combinations of architectural variations, including the presence or absence of Batch Normalization and Skip connections (removing all the skip connections or only the one connected with the output layer).\footnote{The presence of the skip connections could favour a low-level brightness-dependance in the developed weights, fostering trivial solutions in datasets composed by simple patterns and recognizable colors. We explore the removal of such connection following these intuitions. }
%and a sigmoidal activation function on the output layer.
%Moreover, we denote with \textsc{FCN-ND} a Fully-Convolutional model that maintains the same image-resolution throughout the whole architecture, without any downsampling \cite{sherrah2016fully,yu2015multi}. 
%It is based on 5x5 filters padded by 2 pixels, with six layers composed by 32 filters in each hidden layer -- except from the first one, which contains 16 filter banks. 
We compared the obtained features against those produced by massively pretrained state-of-the-art models in Semantic Segmentation. We considered the Dense Prediction Transformer (\textsc{DPT}) \cite{DBLP:journals/corr/abs-2103-13413} and \textsc{DeepLabV3} \cite{chen2017deeplab} with ResNet101 backbone, exploiting both the features produced by the penultimate layer in the classification heads (\textsc{-C} suffix) and the ones obtained by the backbones (i.e., upsampling the representations if needed, \textsc{-B} suffix).
In this way, we investigate both lower-level features based on backbones pretrained on millions of images (ImageNet), and task-specialized higher-level features for semantic segmentation (COCO \cite{lin2014microsoft} and ADE20k \cite{zhou2019semantic} datasets---the latter explicitly includes the categories of the considered textured objects). As \textsc{Baseline} model we considered the case in which the pixel representations are left untouched (i.e., pixel color/brightness).
%Finally, we noticed that the low-level brightness of each channel of the input space (BW or RGB color) represent informative features on their own in the task at-hand,  given the fact that simple objects tend to be characterized by a common color.
%Hence, we simply pick such brightness features ($p = 3$, composed by the repeated gray channel in the \textsc{BW} case) as a baseline, denoted with \textsc{Identity}.

%\vspace{-3mm}
\vspace{-1mm}\myparagraph{Parameters.} Parameters of the attention model were either fixed ($\alpha_m = 1$), or adapted according to a preliminary run of the model ($\alpha_b$, $\rho$). For each video stream, we searched for the model hyper-parameters that maximize the F1 along the attention trajectory, measured during the $30$-th laps (hyper-params grids and the best selected are reported in the suppl. material).
% following grids
% : $\alpha \in \{ 5\cdot10^{-6}, 10^{-4}, 5\cdot10^{-4}, 10^{-3}\}$, $\lambda_T \in \{10^{-2}, 10^{-1}, 1 \}$, $\lambda_S \in \{10^{-4}, 10^{-3},  10^{-2} \}$, $\lambda_C \in \{10^{-4}, 10^{-2}, 10^{-1}, 1, 10\}$, $d \in \{ 32, 128\}$, $e \in \{5\,\text{k}, 10\,\text{k}, 30\,\text{k}\}$, $\beta \in \{3, 5\}$.\footnote{We also evaluated the possibility of removing the outer skip connections in \textsc{HourGlass}, and all the models tested normalized feature vectors (unitary norm) and not normalized ones.} 
% For simplicity, we assumed the model to be aware of the optimal open-set threshold $\xi$, that we found by searching on a dense grid. 
Fig.~\ref{fig:streams} (bottom) shows the effects of the parameters $e$ and $\beta$ in modeling the stochastic coherence graph.

%Finally, we compare the case of feature vectors with unitary Euclidean norm with the unnormalized ones. 
%Depending on the choice, we choose the best performances on a list of  $\xi$ values for the open set classifier $c(\cdot, \xi, \zeta)$. (come inserire le liste del commento??)

% dist_threshold = [0.000001, 0.0005, 0.0003, 0.0002, 0.0007, 0.001, 0.01, 0.1, 0.25, 0.5, 0.7, 1.0 ]
%    dist_threshold = [0.1, 2, 10, 18, 25, 50, 75, 100, 125, 150, 175, 200, 250, 300, 400, 500, 600]

%Overall, we selected the parameters returning the best F1 score when measured on the FOA trajectory in the considered temporal window.

%open set -> background

\setlength{\fboxsep}{0pt}%
\setlength{\fboxrule}{1pt}%

\iffalse
\begin{figure}
    \centering
    \fbox{\includegraphics[width=.19\textwidth]{fig/sampling/original_pairs_1000_spread_05.pdf}}
    \fbox{\includegraphics[width=.19\textwidth]{fig/sampling/plot_sampling_pairs_1000_spread_05.pdf}}
    % \fbox{\includegraphics[width=.24\textwidth]{fig/sampling/plot_sampling_pairs_1000_spread_5.pdf}}
     \fbox{\includegraphics[width=.19\textwidth]{fig/sampling/original_pairs_20000_spread_1.pdf}}
    \fbox{\includegraphics[width=.19\textwidth]{fig/sampling/plot_sampling_pairs_20000_spread_1.pdf}}
    \fbox{\includegraphics[width=.19\textwidth]{fig/sampling/plot_sampling_pairs_20000_spread_5.pdf}}
    \caption{Node sampling in the \textit{stochastic spatial attention graph}  $\tilde{\mathcal{G}}_t$. Left-to-right, a frame sampled from the \textsc{EmptySpace} stream where a laptop is moving, followed by the sampled $\tilde{\mathcal{G}}_t$ with $e=1000$ and $\beta=1.0$. Red nodes represent the FOA coordinates $a_t$, while yellow nodes are sampled in the moving region containing $a_t$. Orange nodes are sampled outside the focused region. In the third figure an ewer is moving. In 4th and 5th figures, the sampling procedure with $e=20k$, first with $\beta=1.0$ and then $\beta=5.0$, respectively.  }
    \label{fig:spread}
\end{figure}
\fi

\begin{figure}
    \tiny
    \centering
     \fbox{\includegraphics[width=.13\textwidth]{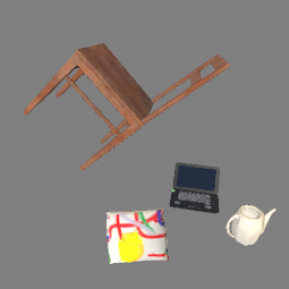}}
     \fbox{\includegraphics[width=.13\textwidth]{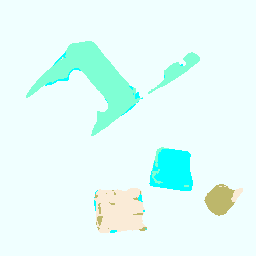}}
     \fbox{\includegraphics[width=.13\textwidth]{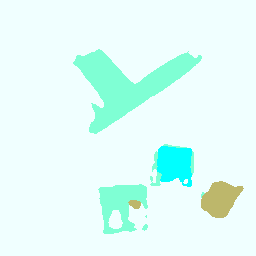}}\\
     \fbox{\includegraphics[width=.13\textwidth]{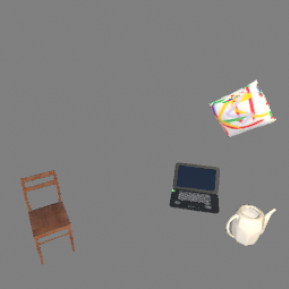}}
     \fbox{\includegraphics[width=.13\textwidth]{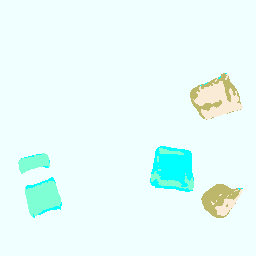}}
     \fbox{\includegraphics[width=.13\textwidth]{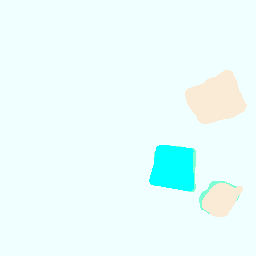}}\\
    \hskip 0.2cm Frame \hskip 1.6cm \textsc{HourGlass} \hskip 1.6cm  \textsc{DPT-B}
    %\vskip -.4cm
    %\hskip 0.9cm Frame \hskip 1.4cm \textsc{HourGlass} \hskip 1.1cm  \textsc{DPT-B}  %\hskip 1.7cm Frame \hskip 1.4cm \textsc{HourGlass} \hskip 1.2cm  \textsc{DPT-B} 
    \caption{Predictions in two frames, comparing \textsc{HourGlass} (our) with \textsc{DPT-B} (transformers) in \textsc{EmptySpace} stream. Different colors indicate different predictions.}
    \label{fig:predictions}
    \vskip -2mm
\end{figure}

%\vspace{-3mm}
\myparagraph{Results.} 
Table~\ref{tab:main_foa} (top) shows the F1 along the attention trajectory measured during the latest object lap, averaged over 3 runs with different initialization. The proposed learning mechanism is competitive and able to overcome models pretrained on large supervised data. This is mostly evident in the case of \textsc{HourGlass}, while \textsc{FCN-ND} is less accurate, mostly due to the lack of implicit spatial aggregation.
%Results show that what we propose enables a plausible attention-based interaction with the human, since the agent could be potentially able to ``speak'' about what it observes.
Table~\ref{tab:main_foa} (bottom) is about the F1 computed considering predictions on all the pixels of all the frames. This measure is not directly affected by the attention model, and while our model performs worse, it is still on par with some of the competitors, with the exception of  \textsc{LivingRoom} (RGB). 
\setlength{\tabcolsep}{2pt}
\begin{table}[t]
\centering
% \scalebox{0.9}{
\footnotesize
\begin{tabular}{c|ccccc}
\toprule
 & \multicolumn{2}{c}{\textsc{EmptySpace}} &  \textsc{Solid}  & \multicolumn{2}{c}{\textsc{LivingRoom}} \\ 
\cmidrule(l){2-3}  \cmidrule(l){4-4}    \cmidrule(l){5-6}   \\
 [-5mm] & {\tiny BW} & {\tiny RGB} & {\tiny BW} & {\tiny BW} & {\tiny RGB} \\
\toprule
\textsc{DPT-C}&$0.70$&$0.76$&$0.54$&$0.38$&$0.39$\\
\textsc{DPT-B}&$\textbf{0.73}$&$\textbf{0.77}$&$0.53$&$0.29$&$0.32$\\
\textsc{DeepLab-C}&$0.46$&$0.57$&$0.56$&$0.19$&$0.26$\\
\textsc{DeepLab-B}&$0.61$&$0.72$&$0.66$&$0.28$&$0.35$\\
\textsc{Baseline}&$0.39$&$0.61$&$0.32$&$0.52$&$\textbf{ 0.65 } $\\
\midrule
\textsc{HourGlass}&$\textbf{0.73}$\scalebox{.7}{$\pm 0.03$}&$\textbf{0.77}$\scalebox{.7}{$\pm 0.05$}&$\textbf{0.68}$\scalebox{.7}{$\pm 0.02$}&$\textbf{0.59}$\scalebox{.7}{$\pm 0.02$}&$0.45$\scalebox{.7}{$\pm 0.12$}\\
\textsc{FCN-ND}&$0.69$\scalebox{.7}{$\pm 0.03$}&$0.57$\scalebox{.7}{$\pm 0.08$}&$0.58$\scalebox{.7}{$\pm 0.07$}&$0.39$\scalebox{.7}{$\pm 0.02$}&$0.38$\scalebox{.7}{$\pm 0.02$}\\
\bottomrule 
%\end{tabular}
% }
\multicolumn{6}{c}{$\ $} \\
\toprule
\textsc{DPT-C}&$0.66$&$0.67$&$0.64$&$0.35$&$0.39$\\
\textsc{DPT-B}&$\textbf{ 0.71 }$&$0.69$&$\textbf{ 0.68 } $&$\textbf{ 0.39 } $&$0.39$\\
\textsc{DeepLab-C}&$0.49$&$0.61$&$0.57$&$0.31$&$0.34$\\
\textsc{DeepLab-B}&$0.70$&$0.65$&$0.66$&$0.34$&$\textbf{0.44} $\\
\textsc{Baseline}&$0.50$&$0.45$&$0.18$&$0.10$&$0.23$\\
\midrule
\textsc{HourGlass}&$0.55$\scalebox{.7}{$\pm 0.03$}&$\textbf{0.71}$\scalebox{.7}{$\pm 0.03$}&$0.50$\scalebox{.7}{$\pm 0.01$}&$0.31$\scalebox{.7}{$\pm 0.04$}&$0.25$\scalebox{.7}{$\pm 0.07$}\\
\textsc{FCN-ND}&$0.60$\scalebox{.7}{$\pm 0.05$}&$0.51$\scalebox{.7}{$\pm 0.07$}&$0.48$\scalebox{.7}{$\pm 0.03$}&{$0.24$\scalebox{.7}{$\pm 0.01$}}&{$0.28$\scalebox{.7}{$\pm 0.03$}} \\
\bottomrule 
\end{tabular}
\caption{Top: F1 score (mean $\pm$ std) measured along the attention trajectory. Competitors are not affected by the model initialization (no std). Bottom: F1 considering all the pixels.}\label{tab:main_foa}
\end{table} 
We investigated our results, showing in Fig.~\ref{fig:predictions} some sample predictions comparing \textsc{HourGlass} with transformers (\textsc{DPT-B}). Indeed, state-of-the-art models have troubles in recognizing closer objects and also in discriminating from the background, due to the fact that their pixel representations are typically strongly affected by a large context. 
% These models expect a classifier that further elaborates the information encoded in pixels belonging to local areas, 
% trained in a fully supervised manner that is not realistic in our setting.
When presenting objects in unusual orientations, likely different from what observed during the fully supervised training, they tend to perform badly. Differently, our model adapts to the video stream, learning more coherent representations when the object transforms. Overall, the attention model allows to focus on what is more important, and, although just a tiny number of supervisions are provided, our model can online-learn to make predictions that competes with the massively offline trained competitors.

\begin{figure}[!t]
\centering
\includegraphics[height=2.49cm]{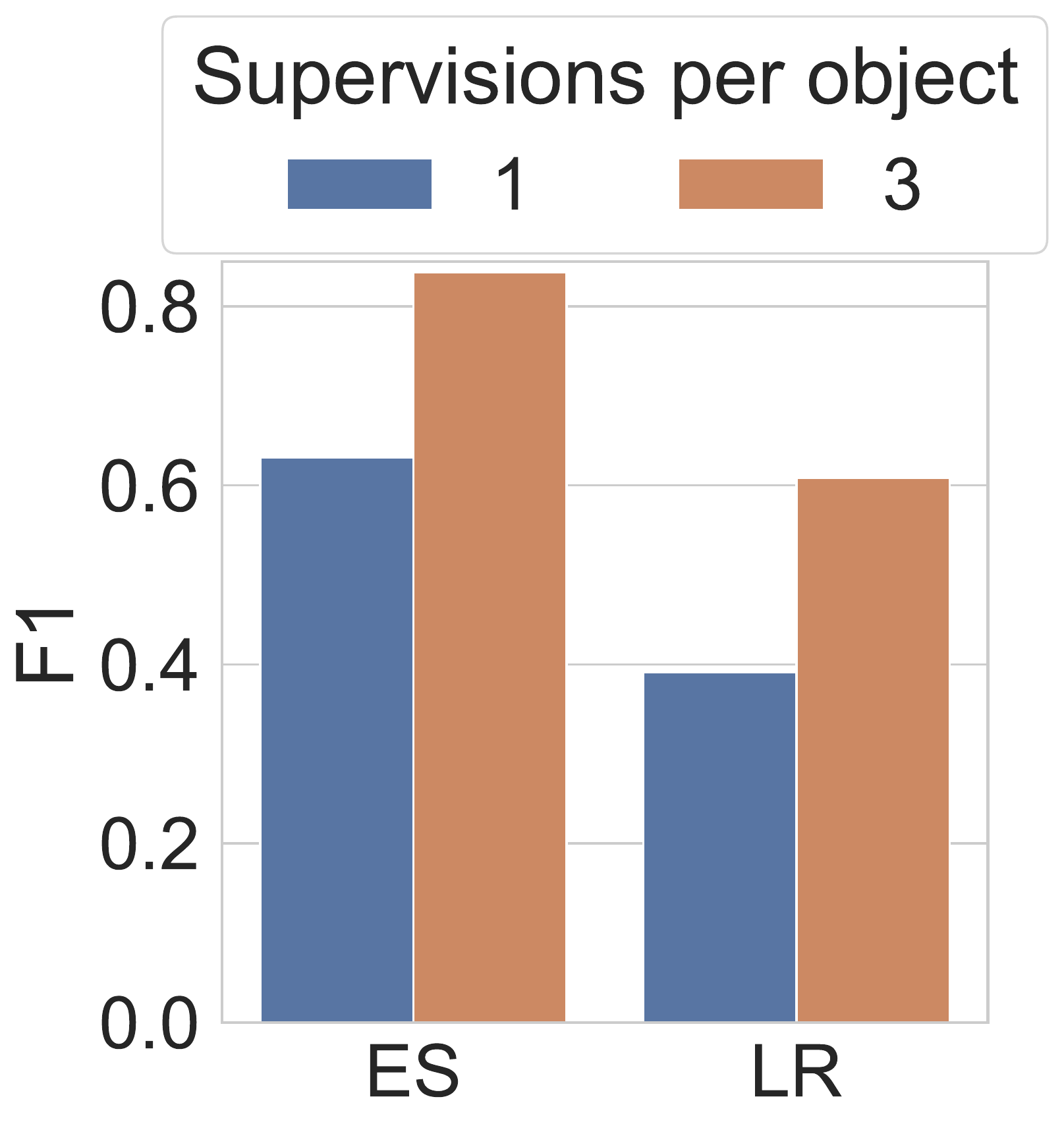}
\includegraphics[height=2.49cm,trim=45 0 5 0,clip]{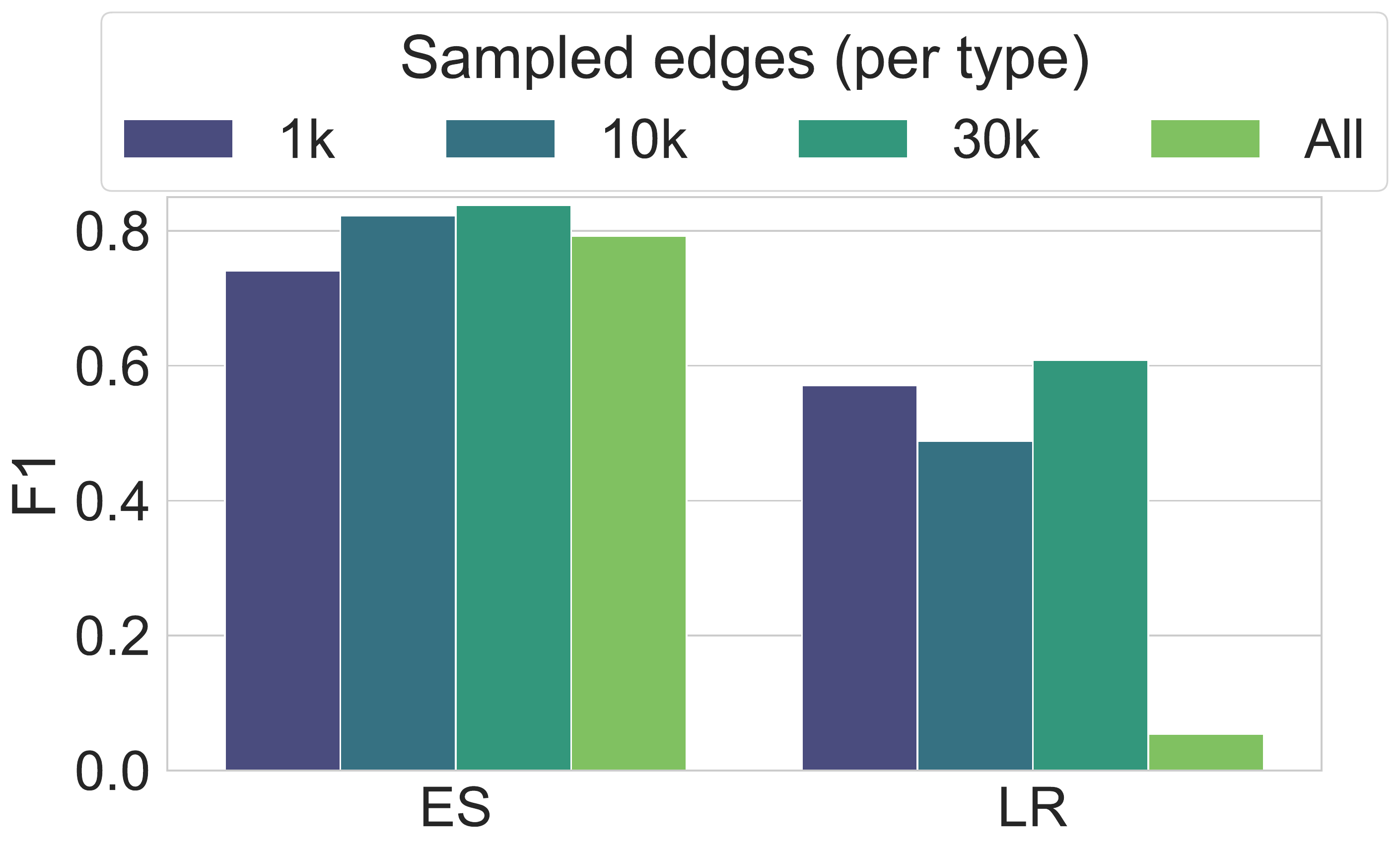}
\includegraphics[height=2.49cm,trim=45 0 0 0,clip]{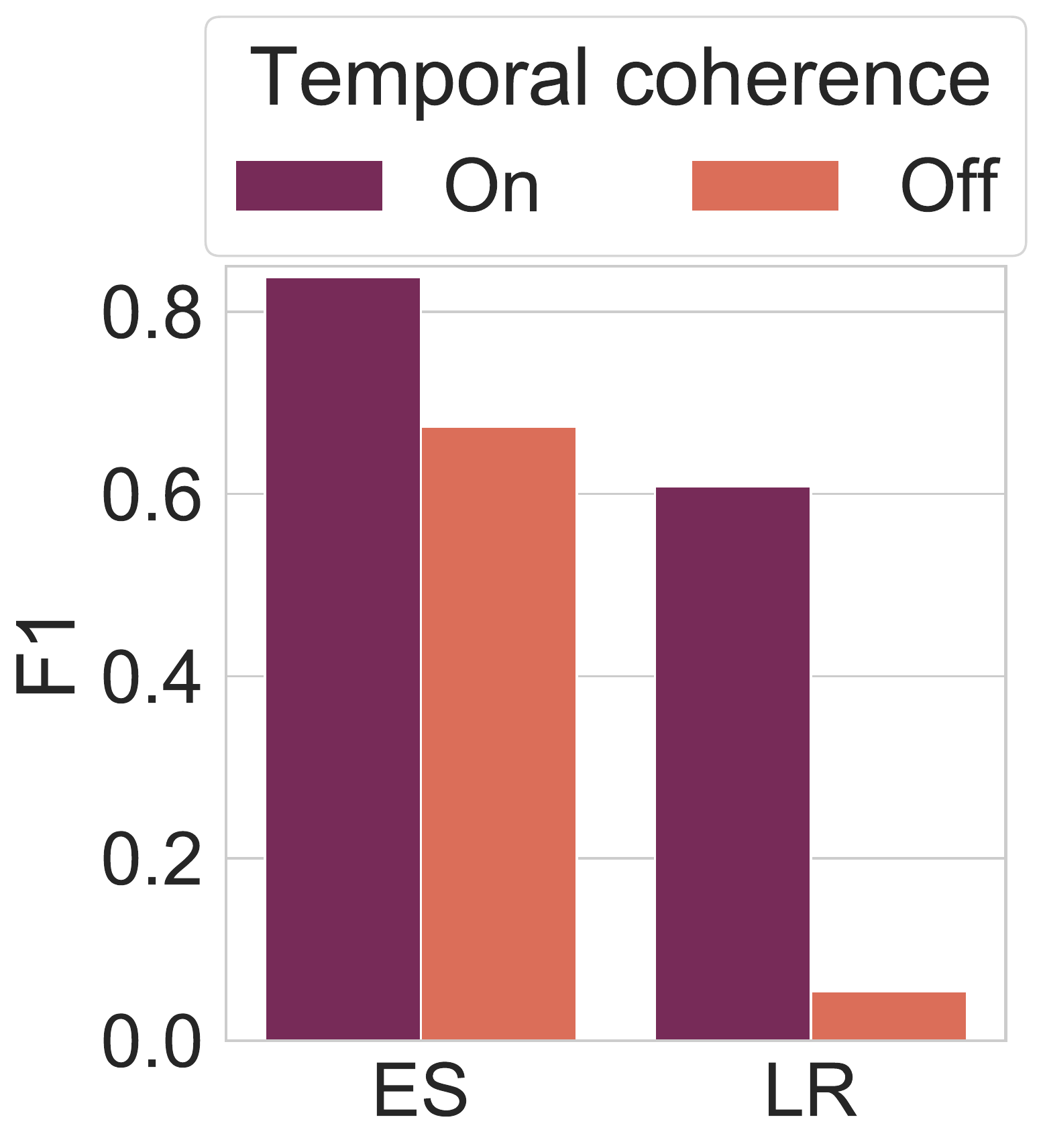}\\
\includegraphics[height=2.49cm,trim=45 0 0 0,clip]{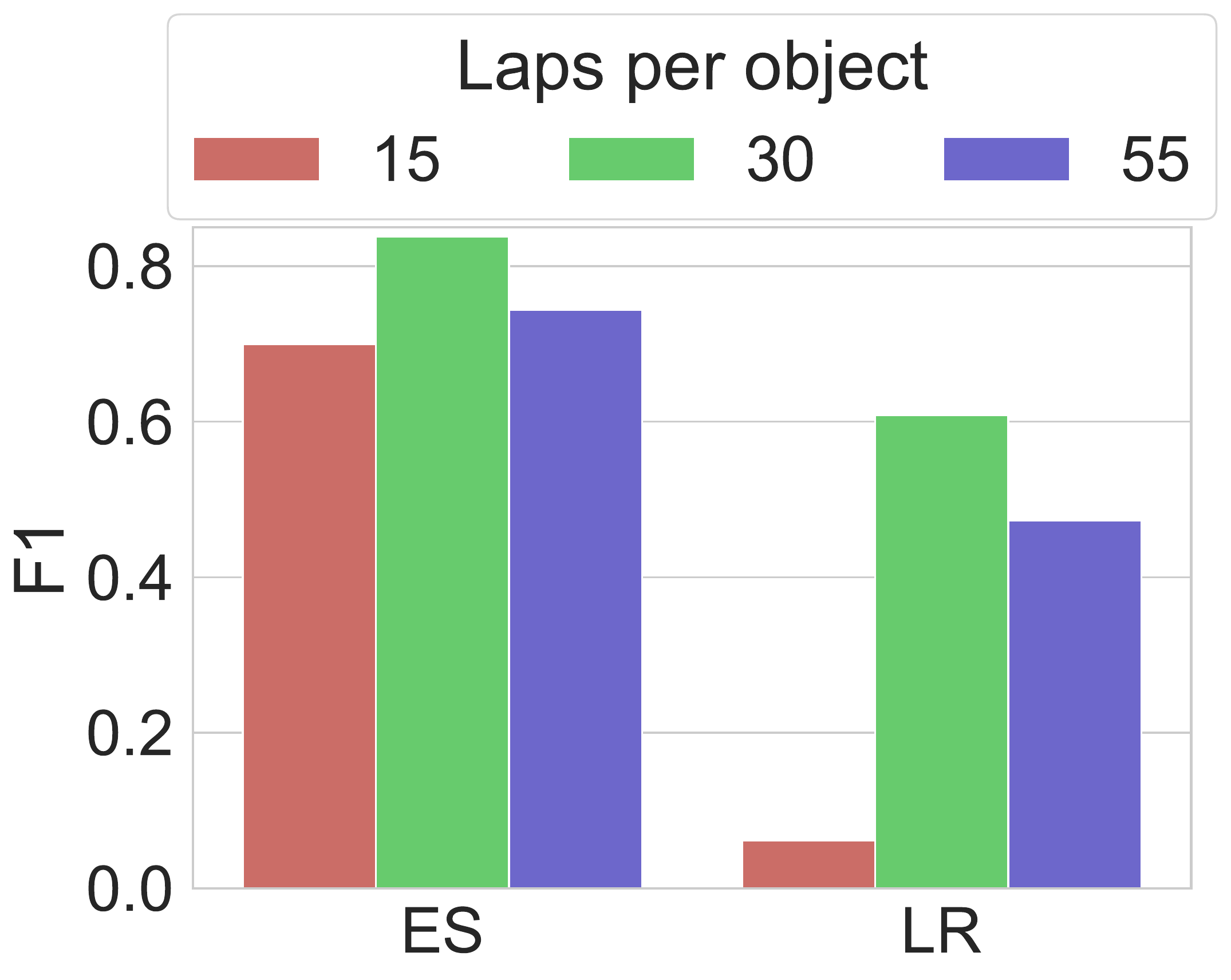}
\includegraphics[height=2.49cm,trim=8 0 0 0,clip]{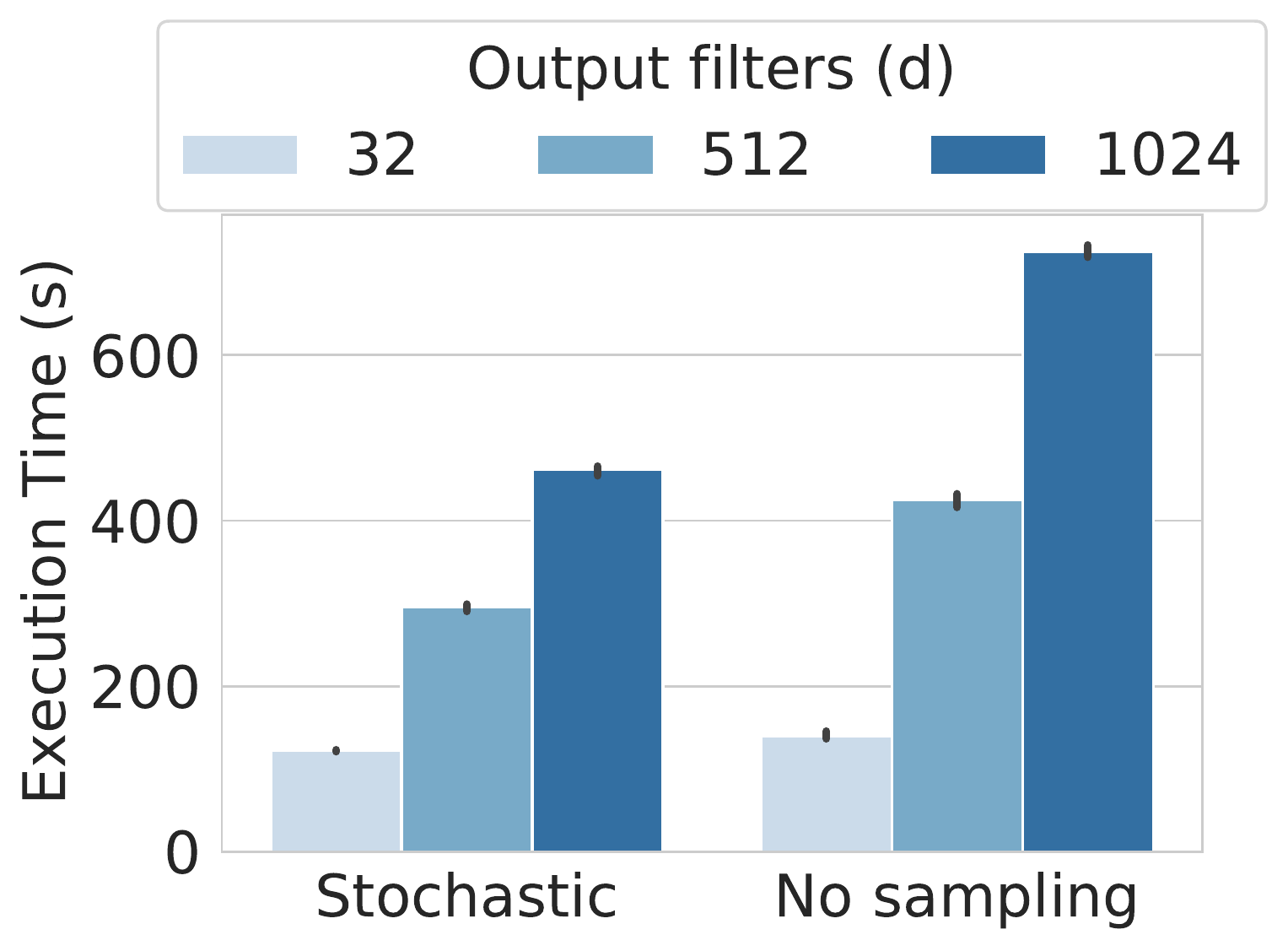}
\vskip -1mm
\caption{First 4 plots (left-to-right, top-to-bottom): in-depth experiments on \textsc{EmptySpace} (ES) and \textsc{LivingRoom} (LR) RGB streams (F1 on focus trajectory). Last plot: timings (3 runs) for stochastic sampling ($e=10\,$k) and no subsampling at all, varying $d$.}
\label{fig:indepth-all}
\vskip -1mm
\end{figure}

\myparagraph{In-depth studies and ablations.}
In order to evaluate the sensitivity of the proposed approach to the key elements of the considered setup, we selected the \textsc{HourGlass} model of Table~\ref{tab:main_foa}. Fig.~\ref{fig:indepth-all} reports results of experiments in which we changed the number of supervisions per object, the number of edges  $e$ (per type) in the stochastic graph, we disabled the temporal coherence, and we changed the length of the streams (discarding \textsc{Solid} in which differences were less appreciable).
Even with a single supervision, the model is able to distinguish the target objects in \textsc{EmptySpace}, while in the more cluttered \textsc{LivingRoom} it benefits from multiple supervisions, as expected. Our proposal is better than using a non-stochastic criterion  (\textsc{LivingRoom}), and works well even with a limited value for $e$. Moreover, the agent benefits from relatively longer streams (going from 15 to 30 laps), that allow it to develop more coherent representations (we tuned the model in the 30-lap case, that is the reason for the slight performance drop in 55-lap), and temporal coherence has an important role in the overall results.
For completeness, we highlighted in Fig.~\ref{fig:indepth-all} (last) the computational benefits, in terms of time (one lap per object), brought by the stochastic subsampling, showing that they are more evident for large $d$. %We report the execution time of the \textsc{HourGlass} model on one complete \textit{route} per object in the \textsc{EmptySpace} stream, comparing the case of subsampling $e=10k$ edges, and the case of considering all the possible edges, varying the  output features dimension $d$. 

\section{Conclusions}
\label{sec:conclusions}
% 
% We presented a novel approach to the design of agents that learn in an online manner in a visual environment,  based on a focus of attention mechanism and spatio-temporal coherence.  The proposed agents autonomously develop features useful in distinguishing moving objects, in an open-set class-incremental setting, following an innovative way of benchmarking this class of algorithms using 3D Virtual Environments.

We presented a novel approach to the design of agents that continuously learn from a visual environment dealing with an open-set class-incremental setting, leveraging a focus of attention mechanism and spatio-temporal coherence. We devised an innovative way of benchmarking this class of algorithms using 3D Virtual Environments, in which our proposal leads to results that, on average, are comparable to those obtained using state-of-the art strongly-supervised models. 
% , paving the road to the exploration of attention-based linguistic interaction with humans from virtual to real-world environments. 

% Learning representations from scratch in an online manner leads to results that, on average, are comparable to those obtained using state-of-the art strongly-supervised pretrained models, paving the road to the exploration of attention-based linguistic interaction with humans from virtual to real-world environments. 
%From a very generic perspective, agents that continuously learn can have a very interesting technological impact, but they can also be interpreted as tools to create uncontrolled autonomous systems that might act in a malicious manner. However, this work focusses on very specific aspects that link vision and time, and that do not yield any evident potential negative societal impacts. Moreover, in a longer-term perspective, what we propose can overcome privacy and security-related issues, since our agents can fully learn without using models pretrained on eventually uncontrolled databases, living in a private target environment.
%More work is needed to reach the level in which the proposed model is fully robust and ready to be moved to real-world cases, that will be the outcome of a long term process. 
% This work opens the door to the exploration of attention-based linguistic interaction with humans, that will be our future work, moving from virtual to real-world environments. 

% \clearpage 

\section*{Acknowledgements}
This work was partly supported by the PRIN 2017 project RexLearn, funded by the Italian Ministry of Education, University and Research (grant no. 2017TWNMH2).

\bibliographystyle{named}
\bibliography{biblio}

% \clearpage

\appendix

\begin{center}
\textbf{\Large Supplemental Material}\\
$\ $\\
%\textbf{\Large Stochastic Coherence Over Attention Trajectory For Continuous Learning In Video Streams}
\end{center}

This supplemental paper includes more details about some aspects of the main paper. Moreover, it provides full references to our code, to the instructions to reproduce our results (including all the selected hyperparameters), and, more importantly, to the 3D scenes that we created for this activity. Such scenes are plugged into the 3D Virtual Environment named SAILenv \cite{DBLP:conf/icpr/MeloniPTGM20}, for which we provide further references and instructions. The same scenes were also pre-rendered, saving frames, pixel-wise labels, and optical flow to disk---we are also sharing these data.

\textit{Notice that the bibliographic references in this supplemental paper are about the bibliography of the main paper.}

\section{Normalized Representations}
In addition to what has been described in the main paper, we also explored the case in which the feature vectors are normalized in order to have unitary Euclidean norms,
\begin{equation}
    \| f_x(V_t, \omega_t) \| = 1,\quad \forall x \in Z^\diamond,\quad \forall t\ge0.
    \label{eq:unitary}
\end{equation}
This avoids the system to encode information in the length of the feature vectors, and it can be used to simplify the loss terms of Eq.~\ref{eq:temporal_coherence}, Eq.~\ref{eq:spatial_coherence}, Eq.~\ref{eq:contrastive} as follows,
\begin{eqnarray}
    L_{T}(\omega, \hat{\omega}, t) \hskip -2mm  &:=&  \hskip -2mm \delta_t \left( 1 - \langle f_{a_t}(V_t, \omega), f_{a_{t-1}}(V_{t-1}, \hat{\omega}) \rangle \right) \\
    L_{S}(\omega, t)  \hskip -2mm &:=&  \hskip -2mm \frac{1}{2} \sum_{\substack{x,z \in S_t\\ x \neq z}} \left( 1 - \langle f_{x}(V_t, \omega), f_{z}(V_t, \omega) \rangle \right)\\
    L_{C}(\omega, t)  \hskip -2mm  &:=&  \hskip -3mm \sum_{x \in S_t} \sum_{z \in O_t} \left( 1 + \langle f_{x}(V_t, \omega), f_{z}(V_t, \omega) \rangle \right)
    \label{eq:unitary_norm_losses}
\end{eqnarray}
where $\langle \cdot, \cdot \rangle$ is the dot product and each loss has zero as minimum value. In the case of pixel-wise classification, when using normalized representations we computed the distance between a template $k_t$ and a certain feature vector $f_x(V_{t'},\omega_{t'})$ (with $t' \geq t$) as
\begin{equation}
    \texttt{dist}\left(k_t, f_x(V_{t'},\omega_{t'})\right) = 1 - \langle k_t, f_x(V_{t'},\omega_{t'}) \rangle ,
    \label{eq:dotclass}
\end{equation}
and the open-set threshold $\xi$ belongs to $(0,2]$.

\section{Segmenting the Attended Moving Region}
The coordinates belonging to set $S_t$ are those of the pixels that belong to the moving region that includes the attention coordinates $a_t$. If $v_{x,t}$ is the 2D velocity of pixel at coordinates $x$ and time $t$, then such pixel is a \textit{moving pixel} if and only if it is associated to a non-vanishing optical flow, i.e.,
\begin{equation}
    \|v_{x,t}\| > \gamma,
    \label{eq:segment_thres}
\end{equation}
given $\gamma > 0$ and being $\| \cdot \|$ the Euclidean norm. Two \textit{moving pixels} that are direct neighbors in the image plane are defined as \textit{connected}, and a chain of connected pixels implements what we will refer to as a \textit{path}.
We introduce the function $\texttt{path\_exists}(x,a_t)$, that returns true if there exists a path connecting $x$ to the attention $a_t$.
Hence,
\begin{equation}
S_t = \{x\colon \texttt{path\_exists}(x,a_t) \} \cup \{a_t\} . 
    \label{eq:S}
\end{equation}
If there are no moving pixels in the current frame, then we consider $S_t$ to be empty (in this case, also $a_t \notin S_t$), and the spatial coherence and contrastive losses are set to zero.

In order to segment the moving area as defined above, we implemented a frontier-based algorithm reported in Alg.~\ref{alg:segment}, where the function $\texttt{neighbors(x)}$ returns the set of direct neighbors of $x$.
\begin{algorithm}
 \caption{Determining the moving area that includes the focus of attention.}
 \begin{algorithmic}[1]
 \renewcommand{\algorithmicrequire}{\textbf{Input:}}
 \renewcommand{\algorithmicensure}{\textbf{Output:}}
 \REQUIRE Optical flow $\left[ v_{x,t} \right]_{x \in Z^{\diamond}}$, attention coordinates $a_t$, threshold $\gamma > 0$.
 \ENSURE Set of coordinates belonging to the moving area, i.e., $S_t$.
  \STATE $S_t \leftarrow \{a_t\}$ \hfill \COMMENT{\(\triangleright\) initial set}
  \STATE $\mathcal{F} \leftarrow \{a_t\}$ \hfill \COMMENT{\(\triangleright\) initial frontier}
  \WHILE {$\mathcal{F} \neq \emptyset$}
  \STATE $\mathcal{N} \leftarrow \cup_{x \in \mathcal{F}} \texttt{neighbors}(x)$ \hfill \COMMENT{\(\triangleright\) union of the neighbors of the points in the frontier}
  \STATE $\mathcal{F} \leftarrow \{z: z \in \mathcal{N} \land z \notin S_t \land \|v_{z,t}\| > \gamma \}$ \hfill \COMMENT{\(\triangleright\) new frontier: moving pixels in $\mathcal{N}$ (not in $S_t$)}
  \STATE $S_t \leftarrow S_t \cup \mathcal{F}$ \hfill \COMMENT{\(\triangleright\) adding the new frontier to $S_t$}
  \ENDWHILE
  \IF {$|S_t|=1$}
    \STATE $S_t \leftarrow \emptyset$ \hfill \COMMENT{\(\triangleright\) if only $a_t$ belongs to $S_t$, we clear it}
  \ENDIF  
 \RETURN $S_t$ 
 \end{algorithmic}
 \label{alg:segment}
 \end{algorithm}

 \section{Setup}
 
We created three visual streams ($256 \times 256$), both in grayscale (\textsc{BW}) and color (\textsc{RGB})---\textsc{Solid} is \textsc{BW} only---by observing the moving scenes. To accelerate the experimental validation we generated $\approx 51\,$k, $12\,$k and $20\,$k pre-rendered frames from the three aforementioned streams, respectively, that correspond to $31$ completed laps for each object.
%we extrapolated from each stream a video, which was repeated 25 times (yielding 10k, 41k and  16k frames-long videos, for \textsc{Solid}, \textsc{EmptySpace}, \textsc{LivingRoom} respectively)  in order to constitute the learning data on which the unsupervised criterion of stochastic coherence learning described in Section \ref{sec:model} is performed\footnote{Note that we employ motion-related information (optical flow) directly produced from SAILenv. This statistics could be obtained, in real-life videos, from other algorithms \cite{horn1981determining}. Moreover, we assume to deal with the fixed-camera scenario to ease the optical-flow computation. This assumption can be easily removed filtering out the camera movement as a preprocessing step.}. 
The agent learns by watching the first $25$ laps per object and, afterwards, he receives $3$ supervisions $(k_t, y_t)$ per object at different time instants, starting from the first frame of the route and waiting at least $100$ frames before supervising  the same object again (of course, $a_t$ must belong to the supervised object, and we set $b = 1+3n$ to ensure a frequent update of the templates). When all the objects complete $30$ laps, we stopped learning the model parameters and, finally, the last lap was the one on which performance is measured.
\vspace{-1mm}\paragraph{Parameters.} The parameters of the attention model were either fixed ($\alpha_m = 1$), or adapted according to a preliminary run of the model on the video streams ($\alpha_b$, $\rho$). For each video stream, we searched for the model hyper-parameters that maximize the F1 along the attention trajectory, measured during the $30$-th laps, considering the following grids: $\alpha \in \{ 5\cdot10^{-6}, 10^{-4}, 5\cdot10^{-4}, 10^{-3}\}$, $\lambda_T \in \{10^{-2}, 10^{-1}, 1 \}$, $\lambda_S \in \{10^{-4}, 10^{-3},  10^{-2} \}$, $\lambda_C \in \{10^{-4}, 10^{-2}, 10^{-1}, 1, 10\}$, $d \in \{ 32, 128\}$, $e \in \{5\,\text{k}, 10\,\text{k}, 30\,\text{k}\}$, $\beta \in \{3, 5\}$.\footnote{We also evaluated the possibility of removing the outer skip connections in \textsc{HourGlass}, and all the models tested normalized feature vectors (unitary norm) and not normalized ones.} For simplicity, we assumed the model to be aware of the optimal open-set threshold $\xi$, that we found by searching on a dense grid. Fig.~\ref{fig:streams} (bottom) shows the effects of the parameters $e$ and $\beta$ in modeling the stochastic coherence graph.

\section{Neural Architectures}

\vspace{-1mm}\paragraph{Our models.} The experimental campaign was carried out evaluating the performance of the proposed approach using two different types of deep convolutional architectures with $d$ output filters, referred to as \textsc{HourGlass} and \textsc{FCN-ND}, respectively. 
The former is a UNet \cite{ronneberger2015u} architecture\footnote{\url{https://github.com/usuyama/pytorch-unet} (MIT License).} based on a ResNet18 backbone.
We explored several architectural variations for this model, including the removal of Batch Normalization (that in our setting boils down to spatial normalization \footnote{\url{https://pytorch.org/docs/stable/generated/torch.nn.BatchNorm2d.html}} since we process frames one after the other) from the whole structure and by removing some skip connections. Indeed, differently from what commonly happens in supervised Semantic Labeling where a target is defined on each pixel, in our case, due to the unsupervised criterion, the presence of skip connections could induce the model to not consider the inner portions of the network, developing representations that are extremely local in terms of spatial information that they encode.  This fact could entail trivial solutions in datasets composed by simple objects characterized by several pixels with instance-specific colors, making the RGB information an attractive way for representing pixel-wise features.
Hence, we tested some variations of the original implementation, either removing all the skip connections or only the one connecting the input with the output layer. We treated these variants of the \textsc{HourGlass} network as special hyperparameters of the model. 
Differently, the other model we considered, \textsc{FCN-ND}, is a 6-layer Fully-Convolutional model that maintains the same image-resolution throughout the whole architecture, without any downsamplings/poolings, inspired by \cite{sherrah2016fully}. It is based on $5 \times 5$ filters, with 6 layers composed of 32 filters in each hidden layer, except from the first one, which contains $16$ filter banks. 
In the experiments, we report the average results obtained over 3 runs initialized with random seeds (\texttt{torch.manual\_seed} in the PyTorch library, leveraging the \texttt{time} package). We report in Table \ref{tab:best_params} the hyperparameters corresponding to the best models. See the provided code repository for further details (e.g, on the network variations hyperparameters.)

\setlength{\tabcolsep}{2pt}
\begin{table}[t]
\centering
% \scalebox{0.9}{
\footnotesize
\begin{tabular}{lc|c@{\hspace{1mm}}c@{\hspace{1mm}}c@{\hspace{1mm}}c@{\hspace{1mm}}c}
\toprule
&  & \multicolumn{2}{c}{\textsc{EmptySpace}} &  \textsc{Solid}  & \multicolumn{2}{c}{\textsc{LivingRoom}} \\
\cmidrule(l){3-4}  \cmidrule(l){5-5}    \cmidrule(l){6-7} 
& Parameters & {\tiny BW} & {\tiny RGB} & {\tiny BW} & {\tiny BW} & {\tiny RGB} \\
\toprule
\multirow{8}{*}{\rotatebox{90}{\textsc{HourGlass}}} & $\alpha$ & $5 \cdot 10^{-6}$ & $5 \cdot 10^{-4}$ & $10^{-4}$ &$10^{-3}$&$5 \cdot 10^{-4}$ \\
& $\lambda_T$&1 &1&$10^{-1}$&$10^{-1}$&$10^{-1}$\\
& $\lambda_S$& $10^{-4}$ & $10^{-4}$ &$10^{-3}$& $10^{-2}$ &$10^{-3}$\\
& $\lambda_C$&$10^{-1}$ &$10^{-1}$&$10^{-1}$&10&10\\
& $d$&32& 128 &128&32&32\\
& $e$& $30\,\text{k}$& $30\,\text{k}$ & $10\,\text{k}$ & $30\,\text{k}$&$30\,\text{k}$\\
& $\beta$&5 &5 &5& 5 &5\\
& \textsc{Norm.}& yes & yes & yes & yes & yes\\
\midrule
\multirow{8}{*}{\rotatebox{90}{\textsc{FCN-ND}}}&  $\alpha$ &$10^{-3}$& $10^{-3}$&$10^{-3}$&$10^{-4}$&$10^{-3}$\\
&  $\lambda_T$&$10^{-1}$&$10^{-2}$&$10^{-2}$&$10^{-2}$&$10^{-2}$\\
& $\lambda_S$&$10^{-2}$&$10^{-3}$&$10^{-2}$&$10^{-2}$&$10^{-3}$\\
& $\lambda_C$&$10^{-2}$&$10^{-2}$&$10^{-2}$&1&$10^{-1}$\\
& $d$&32&32&32&128&128\\
&  $e$&$10\,\text{k}$&$10\,\text{k}$&$10\,\text{k}$&$30\,\text{k}$&$30\,\text{k}$\\
& $\beta$&5&3&3&3&3\\
& \textsc{Norm.}&no&no&no&yes&yes\\
\bottomrule 
\end{tabular}
\caption{Optimal parameters. Best selected hyperparameters drawn from the grid search described in the main
text (see Section \ref{sec:exp}), both for \textsc{HourGlass} (top) and \textsc{FCN-ND} (bottom) architectures. Additionally, we denote with \textsc{Norm.}~$\in \{yes, no\}$ the choice of processing normalized feature vectors (unitary norm) and not normalized ones, respectively. For further details, see the provided code repository.
 }\label{tab:best_params}
\end{table}

% $\alpha \in \{ 5\cdot 10^{-5}, 10^{-4}, 5 \cdot 10^{-4}, 10^{-3}\}$, $\lambda_T \in \{10^{-2}, 10^{-1}, 1 \}$, $\lambda_S \in \{10^{-4}, 10^{-3},  10^{-2} \}$, $\lambda_C \in \{10^{-4}, 10^{-2}, 1, 10\}$, $d \in \{ 32, 128\}$, $e \in \{5\,\text{k}, 10\,\text{k}, 30\,\text{k}\}$, $\beta \in \{3, 5\}$.

\vspace{-1mm}\paragraph{Competitors.} DPT \cite{DBLP:journals/corr/abs-2103-13413}\footnote{\url{https://github.com/intel-isl/DPT} (MIT license).} is a recently proposed architecture based on  vision transformers as a backbone for dense prediction tasks, characterized by high resolutions processing and global receptive fields over the whole frame. We leveraged a pretrained implementation on the ADE20K dataset \cite{zhou2019semantic}. We exploited the best-performing model of \cite{DBLP:journals/corr/abs-2103-13413}, that the authors named DPT-Hybrid.
Differently, \textsc{DeepLabv3} \cite{chen2017deeplab} is based on multiple atrous convolutions rates in order to capture multi-scale contexts. We employed a model based on a ResNet101 backbone and finetuned on COCO dataset \cite{lin2014microsoft}.\footnote{Available at \url{https://pytorch.org/vision/stable/models.html}.}

\section{Code, 3D Scenes, Video Streams}

% provides full references to our code, to the instructions to reproduce our results (including all the selected hyperparameters), and, more importantly, to the 3D scenes that we created for this activity. Such scenes are plugged into the 3D Virtual Environment named SAILenv \cite{DBLP:conf/icpr/MeloniPTGM20}, for which we provide further references and instructions. The same scenes were also pre-rendered, saving frames, pixel-wise labels, and optical flow to disk---we are also sharing these data.

This section contains full references to our code and the instructions to reproduce our results.
At the following link: \url{https://github.com/sailab-code/cl_stochastic_coherence} the reader can find all that is needed to run the experiments. We include the code (PyTorch–Modified BSD license), data (MIT license) and the selected hyper-parameters. See the \textsc{README} file for further details on the archive contents and how to execute the code and on the mapping of the parameters name in the code -- see the "Argument description" in HOW TO RUN AN EXPERIMENT. Notice that the \texttt{reproduce\_runs.txt}  contains the command lines (hence, the best selected parameters) required to reproduce the experiments of the main results (also reported in Table \ref{tab:best_params} for completeness).

Moreover, we provide in the same repository (folder \texttt{3denv}) the three visual scenes we created (\textsc{EmptySpace}, \textsc{Solid}, \textsc{LivingRoom}) for this activity, leveraging the 3D Virtual Environment named SAILenv \cite{DBLP:conf/icpr/MeloniPTGM20}. In addition, the same scenes were pre-rendered into the three visual streams we exploited in the experimental section, saving  frames, pixel-wise labels, and motion information to disk. 

In the following, we will describe how to load such scenes into the SAILenv environment and the pre-rendered \texttt{data} folder structure.

\vspace{-1mm}\paragraph{Computing infrastructure.} 
The experimental campaign was mainly carried out on a machine equipped with Ubuntu 18.04.3 LTS, two NVIDIA Tesla V100-SXM2-32GB GPUs, Intel Xeon Silver 4110 (32 cores) @ 2.101GHz CPU, 48 GB of RAM.
Additionally, a portion of the experiments was carried out on two other machines: the former one is equipped with Ubuntu 18.04.1 LTS, three NVIDIA 1080Ti-11GB GPUs, Intel  i9-7900X (20 cores) @ 3.30GHz CPU, 128 GB of RAM.  The latter one with  Ubuntu 18.04.4 LTS, an NVIDIA GTX TITAN-6GB GPU,  Intel  i7-3970X (12 cores) @ 3.50GHz CPU, 64 GB of RAM.

We used \textrm{Numpy} (v1.19.5) for saving numerical arrays and We exploited the \textrm{wandb} platform to organize and track our experiments.

% {\color{red} This section is fully TODO!}
% \begin{itemize}
    % \item Link to the CODE (and DATA) (Google Drive - zip archive). Include all the scripts to run the experiments. The zip must include README (detailed, with map of the parameter names), LICENSE (MIT License - do not change this). A few words about this link.
\subsection{3D scenes}

\begin{figure*}
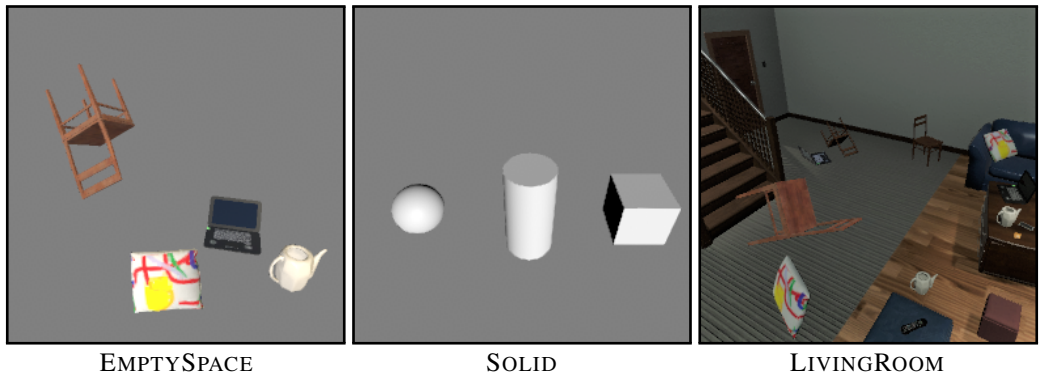

    \centering
    %\begin{minipage}{0.48\textwidth}
    
    % {\centering{\rotatebox[origin=bl]{90}{\textsc{EmptySpace}}}}
    \fbox{\includegraphics[width=.25\textwidth]{fig/streams/empty09_100.png}}
    %\hskip 1mm 
    % {\rotatebox{90}{\textsc{Solid}}}
    \fbox{\includegraphics[width=.25\textwidth]{fig/streams/solid01_001.png}}
    %\hskip 1mm \\
    % {\rotatebox{90}{\textsc{LivingRoom}}}
    \fbox{\includegraphics[width=.25\textwidth]{fig/streams/toy01_065_l.png}}\\
    %\vskip -3.5mm
    % \tiny
   \textsc{EmptySpace} \hskip 3.cm \textsc{Solid} \hskip 3.cm  \textsc{LivingRoom}
 
    %\end{minipage}
    %\hskip 1mm
    %\begin{minipage}{0.44\textwidth}
    %\centering
   
    %\end{minipage}
    \caption{Sample frames from the three 3D scenes/streams (objects rotate and scale while moving). We report the same figures of the main paper (see Figure \ref{fig:streams}, top) in a bigger resolution, to better appreciate the visual stream details.}
    \label{fig:streams_sup}

\end{figure*}

The scenes are available in the \texttt{3denv} folder of the same repository (link \url{https://github.com/sailab-code/cl_stochastic_coherence}). The \texttt{3denv} folder  contains three \texttt{.unity} files and three associated directories.
    \begin{itemize}
        \item \texttt{EmptySpace.unity}: contains the scene named \textsc{EmptySpace}.
        \item \texttt{Solid.unity}: contains the scene named \textsc{Solid}.
        \item \texttt{LivingRoom.unity}: contains the scene named \textsc{LivingRoom}.
    \end{itemize}
    To install the scenes you must follow these instructions: 
    \begin{enumerate}
        \item Download the latest SAILenv source code from SAILenv site (\url{https://sailab.diism.unisi.it/sailenv/}) and extract the files into a directory. 
        \item Download Unity 2019.4.2f1 from Unity website (\url{https://unity3d.com/get-unity/download}). Using Unity HUB is the easiest way. \textbf{NOTE: to use Unity you will need to create an account with a personal free license (Unity Personal).}
        \item Open the SAILenv directory with Unity. The first time it will take around 15 minutes to fully load. 
        \item Copy the content of the scenes directory (\texttt{3denv/scenes}) into \texttt{Assets/Scenes}
        \item Through the Unity Editor, open the file \texttt{Assets/Settings/AvailableScenesSettings}.
        \item The Inspector will show a list of the available scenes. Change the textbox named \textbf{Size} from 4 to 7. This will create three new empty boxes that can be filled by dragging the scenes objects from the Project view to the scene box in the Inspector and assign names to the scenes in the textbox below (see Figure \ref{fig:unity}) (the scenes will be exposed through the Python API with these names - see the example scripts in the official site (\url{https://sailab.diism.unisi.it/sailenv/}).
        
        \begin{figure}
            \centering
            \includegraphics[width=0.9\linewidth]{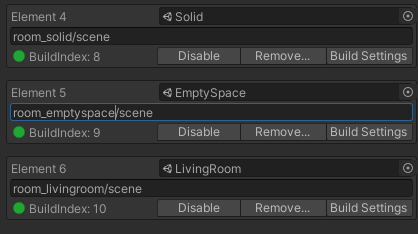}
            \caption{Example on how to assign names to the scenes in the text box in Unity.}
            \label{fig:unity}
        \end{figure}

        \item Press the \texttt{Add (buildIndex N)} and then confirm \texttt{Add as enabled} for each of the newly added scenes.
        \item On the top bar, select \texttt{SAILenv/Builds/Local AssetBundles}. It will take a while to compile the resources.
        \item Open the "Main Menu" scene in \texttt{Assets/Scenes} and press the Play button on the top. There is no need to interact with this scene, just proceed to the next step.
        \item The Environment is now running and you can use the Python API to open the new scenes through the names you have given. 
    \end{enumerate}
The visual stream is made available to clients (default port 8085) by clicking on the play icon in Unity or by building and launching a standalone application. A brief snippet of code that uses the \texttt{InputStream} class (available in our source code) to fetch in real time the rendered stream is available in Figure \ref{fig:access_stream}. We report in Figure \ref{fig:streams_sup} a sample frame from each one of the processed visual streams.

\begin{figure}[h]
    \centering
\begin{minted}[frame=lines,bgcolor=white,fontsize=\codesize,framesep=2mm]{python}
import lve

# sailenv settings for the Solid scene
unity_settings = {
    'depth_frame_active': False,
    'flow_frame_active': True,
    'object_frame_active': False,
    'main_frame_active': True,
    'category_frame_active': True,
    'use_gzip': False,
    'scene': 'room_solid/scene'
}

# create InputStream object
ins = lve.InputStream("localhost:8085", w=256, h=256, 
                        fps=8, repetitions=20, 
                        force_gray=False,
                        unity_settings=unity_settings)
                     
# set agent camera as grey solid background
agent = ins.get_unity_agent()
agent.change_main_camera_clear_flags(128, 128, 128)

# place agent into its default position for the scene
agent.toggle_follow()

# getting next frame,  motion, supervision and foa traj.                     
img, of, supervisions, foa = ins.get_next()  
                          
\end{minted}
    \caption{Code snippet to access the SAILenv scenes  using the Python API and the provided \texttt{InputStream} class. Frame data are obtained with the   \texttt{.get\_next()} method.}
    \label{fig:access_stream}
\end{figure}

\subsection{Video Streams folders}

We rendered the three scenes and extracted the corresponding  visual streams described in Section \ref{sec:exp}. The \texttt{data} folder contains one subfolder for each one of such visual streams. The directory tree of each one of such folders (named after the streams name \textsc{EmptySpace},  \textsc{Solid}, \textsc{LivingRoom}) is composed as follows.

\textbf{\texttt{.foa} file} It is the output of the FOA (focus of attention) computation. The attention is stored in a CSV-like format: \texttt{foa\_x, foa\_y, v\_x, v\_y, saccade}, with Focus of Attention (FOA) frame coordinates and velocity, besides a boolean flag which is true if a saccade is detected for the current frame.

\textbf{ \texttt{frames} directory} It contains the full-resolution frames, stored in PNG format in subfolders of 100 frames.

\textbf{ \texttt{motion} directory} It contains the motion arrays (pixel-wise XY velocity, in $(w, h, 2)$-shaped tensors), stored in NumPy format (gzip compressed, \textrm{.bin} extension) in subfolders of 100 frames.

\textbf{ \texttt{sup} directory} It contains the frame-wise supervision arrays (\texttt{targets} and \texttt{indices}), both stored in NumPy format (gzip compressed, \textrm{.bin} extension) in subfolders of 100 frames. Such supervision is stored in a sparse manner, given the fact that we consider pixel-wise supervisions on the FOA coordinates $a_t$.
% in our work supervision is regarded as a rare action.
\texttt{targets} contains the target class of the supervision for some pixels, while  \texttt{indices} contains the raveled index of the corresponding supervised pixels.

\paragraph{Loading the pre-rendered scenes.} The \texttt{InputStream} class utility can be used to load into Python the prerendered scenes. Figure \ref{fig:input_stream} contains a simple code example to load one of the streams and get some data from it. In particular, the \texttt{get\_next()} method returns the next frame, its optical flow, pixel-wise supervisions and the coordinates attended by the Focus of Attention in that frame.

\begin{figure}[h]
    \centering
\begin{minted}[frame=lines,bgcolor=white,fontsize=\codesize,framesep=2mm]{python}
import lve

# precomputed FOA trajectory file
foa_file="data/emptyspace/empty_space_bench_foa_long.foa"
# create InputStream object
ins = lve.InputStream("data/emptyspace", w=-1, h=-1, 
                        fps=None, max_frames=None,
                        repetitions=1, force_gray="yes", 
                        foa_file=foa_file)
          
# getting next frame,  motion, supervision and foa traj.                      
img, of, supervisions, foa = ins.get_next()  
                          
\end{minted}
    \caption{Loading the pre-rendered \textsc{EmptySpace} visual stream through the \texttt{InputStream} class and getting the data from the next frame using Python.}
    \label{fig:input_stream}
\end{figure}

% {\color{red}    
%  Describe the DATA folder of our video streams (first link above): images, optical flow, etc.. of the 3 VIDEO STREAMS. Some words about the streams, in particular, an example on how to load them with our already working InputStream class.
% }

\end{document}